\documentclass[10pt,twocolumn,letterpaper]{article}
\usepackage{cvpr}              

\usepackage[dvipsnames]{xcolor}
\usepackage[accsupp]{axessibility}
\definecolor{cvprblue}{rgb}{0.21,0.49,0.74}
\usepackage[pagebackref,breaklinks,colorlinks,citecolor=cvprblue]{hyperref}

\def\paperID{5508} 
\def\confName{CVPR}
\def\confYear{2024}

\title{Bilateral Event Mining and Complementary for Event Stream Super-Resolution}

\author{Zhilin Huang$^{1,2}$\thanks{Equal Contribution}\quad Quanmin Liang$^{2,3}$\footnotemark[1]\quad Yijie Yu$^{1,2}$\quad Chujun Qin$^{4}$\quad \\ Xiawu Zheng$^{2,5}$\quad Kai Huang$^{3}$\quad Zikun Zhou$^{2}$\footnotemark[2]\quad Wenming Yang$^{1,2}$\thanks{Corresponding Author}\\
$^1$ Shenzhen International Graduate School, Tsinghua University $^2$ Peng Cheng Laboratory\\
$^3$ School of Computer Science and Engineering, Sun Yat-Sen University\\
$^4$ China Southern Power Grid $^5$ Xiamen University\\
{\tt\small \{zerinhwang03,chujun.qin\}@pku.edu.cn, \tt\small liangqm5@mail2.sysu.edu.cn, \tt\small yangelwm@163.com}\\
{\tt\small yyj23@mails.tsinghua.edu.cn, \tt\small huangk36@mail.sysu.edu.cn, \tt\small \{zhengxw01,zhouzk01\}@pcl.ac.cn}
}

\usepackage{amsmath,amsfonts,bm}

\def\eqref#1{equation~\ref{#1}}

\def\1{\bm{1}}

\def\rve{{\mathbf{e}}}

\def\rvn{{\mathbf{n}}}

\def\rvp{{\mathbf{p}}}

\def\rmA{{\mathbf{A}}}

\def\rmH{{\mathbf{H}}}

\def\rmK{{\mathbf{K}}}

\def\rmQ{{\mathbf{Q}}}

\def\rmV{{\mathbf{V}}}
\def\rmW{{\mathbf{W}}}

\def\rmZ{{\mathbf{Z}}}

\DeclareMathAlphabet{\mathsfit}{\encodingdefault}{\sfdefault}{m}{sl}
\SetMathAlphabet{\mathsfit}{bold}{\encodingdefault}{\sfdefault}{bx}{n}

\newcommand{\R}{\mathbb{R}}

\usepackage{pifont}

\usepackage{stmaryrd}
\usepackage{amsmath}
\usepackage{amssymb}
\usepackage{mathtools}
\usepackage{amsthm}
\usepackage{dsfont}
\usepackage{diagbox}
\usepackage{adjustbox}
\usepackage{multirow}
\usepackage{makecell}
\usepackage{graphicx}
\usepackage{caption}
\usepackage{booktabs}
\usepackage{cellspace}
\usepackage{pifont}
\usepackage{amssymb}
\usepackage{tabularx}

\theoremstyle{plain}

\theoremstyle{definition}

\theoremstyle{remark}

\usepackage{xspace}
\usepackage[capitalize,noabbrev]{cleveref}
\usepackage[textsize=tiny]{todonotes}
\crefname{section}{Sec.}{Secs.}
\Crefname{section}{Section}{Sections}
\Crefname{table}{Table}{Tables}
\crefname{table}{Tab.}{Tabs.}
\usepackage{xspace}
\usepackage{yhmath}
\usepackage{pifont}
\newcommand{\method}{\textsc{BMCNet}\xspace}
\newcommand{\module}{\textsc{BIE}\xspace}

\usepackage{algorithm}
\usepackage{listings}

\definecolor{codegreen}{rgb}{0,0.6,0}
\definecolor{codegray}{rgb}{0.5,0.5,0.5}
\lstdefinestyle{mystyle}{
    commentstyle=\color{codegreen},
    keywordstyle=\color{magenta},
    numberstyle=\tiny\color{codegray},
    stringstyle=\color{codepurple},
    basicstyle=\ttfamily\footnotesize,
    breaklines=true,                 
    keepspaces=true,                 
    numbers=left,                    
    showstringspaces=false,
}
\begin{document}

\maketitle
\begin{abstract}
Event Stream Super-Resolution (ESR) aims to address the challenge of insufficient spatial resolution in event streams, which holds great significance for the application of event cameras in complex scenarios. Previous works for ESR often process positive and negative events in a mixed paradigm. This paradigm limits their ability to effectively model the unique characteristics of each event and mutually refine each other by considering their correlations. In this paper, we propose a bilateral event mining and complementary network (\textbf{BMCNet}) to fully leverage the potential of each event and capture the shared information to complement each other simultaneously. Specifically, we resort to a two-stream network to accomplish comprehensive mining of each type of events individually. To facilitate the exchange of information between two streams, we propose a bilateral information exchange (\textbf{BIE}) module. This module is layer-wisely embedded between two streams, enabling the effective propagation of hierarchical global information while alleviating the impact of invalid information brought by inherent characteristics of events. The experimental results demonstrate that our approach outperforms the previous state-of-the-art methods in ESR, achieving performance improvements of over \textbf{11\%} on both real and synthetic datasets. Moreover, our method significantly enhances the performance of event-based downstream tasks such as object recognition and video reconstruction. Our code is available at \href{https://github.com/Lqm26/BMCNet-ESR}{https://github.com/Lqm26/BMCNet-ESR}.
\end{abstract}
\section{Introduction}
Event cameras are a novel bio-inspired asynchronous sensor \cite{gallego2020event, lichtsteiner2008128}
with advantages such as high dynamic range, high temporal resolution, and low power consumption \cite{gallego2020event, brandli2014240, lichtsteiner2008128}. 
It holds the potential to provide solutions for a wide range of visually challenging scenarios and has already found extensive applications \cite{bardow2016simultaneous, rebecq2019high, tulyakov2022time, taverni2018front, messikommer2023data, cannici2020differentiable, zhou2018semi, liang2023event, xu2020eventcap, li2023sodformer, kim2021n}. However, most existing event cameras still exhibit lower spatial resolution (e.g., DAVIS346 with a resolution of 346\(\times\)260 \cite{taverni2018front}) and are subject to sensor-induced noise. Increasing the spatial resolution at the hardware level inevitably leads to event loss and increased pixel response latency \cite{gehrig2022high, li2021event}. Therefore, there is a need to achieve super-resolution and enhancement of event streams, enabling better adaptability to various complex scenes.

\begin{figure}[!tp]
\centering
\includegraphics[width=\linewidth]{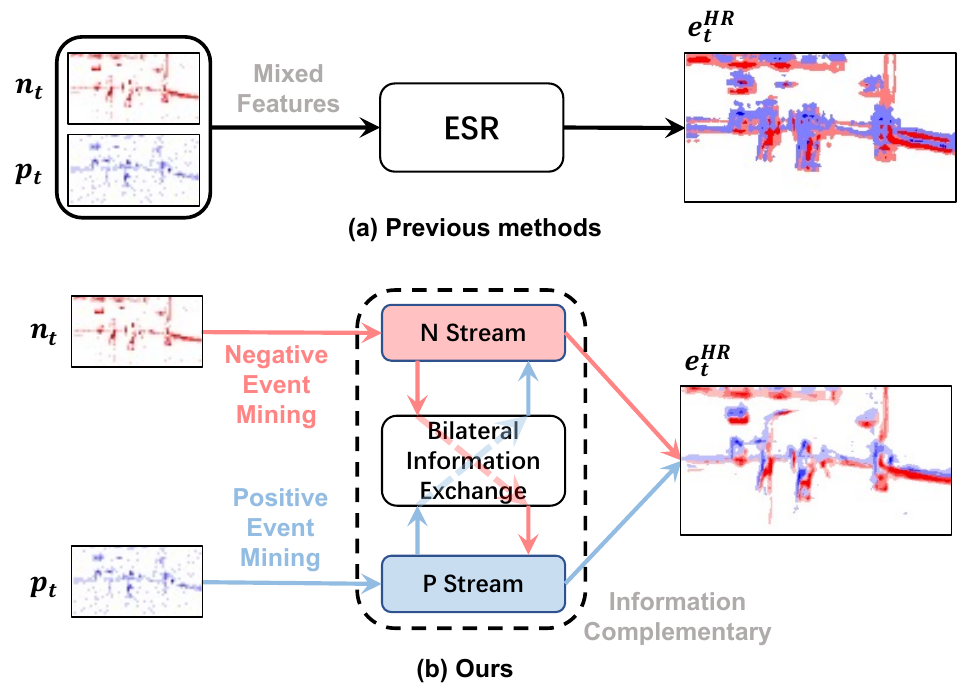}
\caption{Illustration of various approaches in event stream super resolution for processing positive and negative events.}
\vspace{-4mm}
\label{fig:motivation}
\end{figure}

To address this issue, several methods have been proposed for Event Stream Super-Resolution (ESR). One approach involves preserving the spatiotemporal distribution information of event streams, using non-uniform Poisson processes or Spiking Neural Networks (SNNs) to simultaneously estimate the spatiotemporal distribution of event streams during ESR \cite{li2019super, li2021event}. Alternatively, it is possible to employ image frames as auxiliary data, aligning the event stream's resolution with high-resolution images through optical flow estimation \cite{wang2020joint}. However, these approaches face challenges in accurately estimating the spatiotemporal distribution of the event stream, requiring high-quality images as assistance, and struggling to achieve high-factor resolution enhancement. 
Another approach is to apply learning-based methods for super-resolution on event streams \cite{duan2021eventzoom, weng2022boosting}. 
By unifying the data format of events with natural images, these methods can seamlessly leverage the experiences and techniques of well-developed image and video super-resolution methods.
However, all learning-based methods \cite{duan2021eventzoom, weng2022boosting} for ESR ignored explicitly exploring the correlations between two types of events by directly process two different types of events in a mixed way, as shown in \cref{fig:motivation} (a).
Though simple, these approaches require elaborate designs and high capacities in the models, as they need to not only effectively model the distinctive data distribution of each event by addressing inherent characteristics, but also capture the complementary information between two events simultaneously.

In this paper, we propose \textbf{three principles} for the model design of ESR based on inherent characteristics of event data:
(\romannumeral1). Positive and negative events, which are obtained by decoupling the event stream, have unique distribution characteristics. However, they also demonstrate strong correlations in both spatial and temporal domain, allowing them to complement and enhance each other. Therefore, it is crucial for ESR models to effectively mine the information from each type of events and facilitate their interaction.
(\romannumeral2). Events are always triggered near object edges, resulting in event data that primarily consists of global structures. Therefore, the models are required to have the capability to effectively capture and interact global structures between different events. 
(\romannumeral3). Due to the sparsity and noise inherent in events, it is essential to avoid misleading resulting from invalid information during context modeling of event data.

According to these three principles for ESR, we propose a bilateral event mining and complementary network (\method).
In addition to effectively modeling the unique characteristics of each event, \method also possesses the capability to efficiently interact and complement global information between different events.
Specifically, \method consists of two parallel stream for processing positive and negative events, respectively. 
And a novel bilateral information exchange (\module) module we proposed is applied to facilitate the information exchange between two streams.
With a single \module, each channel is treated as a structural representation, and the correlations between two events are modeled across the channel dimension instead of the spatial ones. 
In this way, the global structures of each event can be efficiently captured and transmitted to another event as a complementary, while the potential misleading effects caused by invalid information in the spatial dimensions can be mitigated. 
Additionally, a cross-layer interaction representation (CIR) is introduced into \module for storing useful local and global contexts from previous layers and frames. By layer-wisely stacking the \module, hierarchical information can be propagated between two streams. Moreover, \module could be embedded into each stream and serves as a bridge of exchanging spatial and temporal information of each event. 

The main contributions of our work are as follows:
\begin{itemize}
    \item 
    We propose \method, a two-stream network that models the unique characteristics of positive and negative events in event streams while exploiting complementary spatiotemporal contexts to mutually refine each event other, improving overall performance in ESR.
    
    \item A novel bilateral information exchange (\module) module is proposed to facilitate the exchange of complementary global information between the two types of events while mitigating the impact of invalid information such as noise and empty pixels that are inherent in event data.
    
    \item Experimental results demonstrate that our method achieve a significant improvement of over \textbf{11\%} in ESR, and exhibits a visual fidelity that is closer to that of real event streams.
    
    \item Our approach also effectively enhances downstream tasks, such as event stream recognition and reconstruction, further validating the effectiveness of our method.
\end{itemize}

\section{Related work}
Compared to super-resolution in natural images or videos \cite{chan2021basicvsr, chan2022basicvsr++, gankhuyag2023lightweight, li2019feedback, isobe2022look, geng2022rstt, wang2023compression, chen2023activating, li2023learning, lu2023learning}, ESR presents unique challenges due to its distinct spatiotemporal characteristics. Initial approaches aimed to preserve the spatiotemporal properties of event streams and directly perform super-resolution. Li et al. \cite{li2019super} introduced an Event Count Map (ECM) to describe the spatial distribution of events, modeling events on each pixel using a non-homogeneous Poisson distribution. However, this method often encountered issues of inaccurate spatiotemporal distribution estimation. Subsequently, based on a mixed imaging strategy, Wang et al. \cite{wang2020joint} proposed a novel optimization framework called GEF, which utilizes motion-related probability to filter event noise. With the assistance of image frames, GEF achieves event stream super-resolution. However, GEF may degrade significantly when the quality of image frames diminishes. Li et al. \cite{li2021event}, leveraging the spatiotemporal properties of Spiking Neural Networks (SNNs), introduced a spatiotemporal constraint learning approach capable of simultaneously learning the temporal and spatial features of event streams. Yet, this method falls short of achieving high-factor super-resolution.

\begin{figure*}[!t]
\centering
\includegraphics[width=0.9\linewidth]{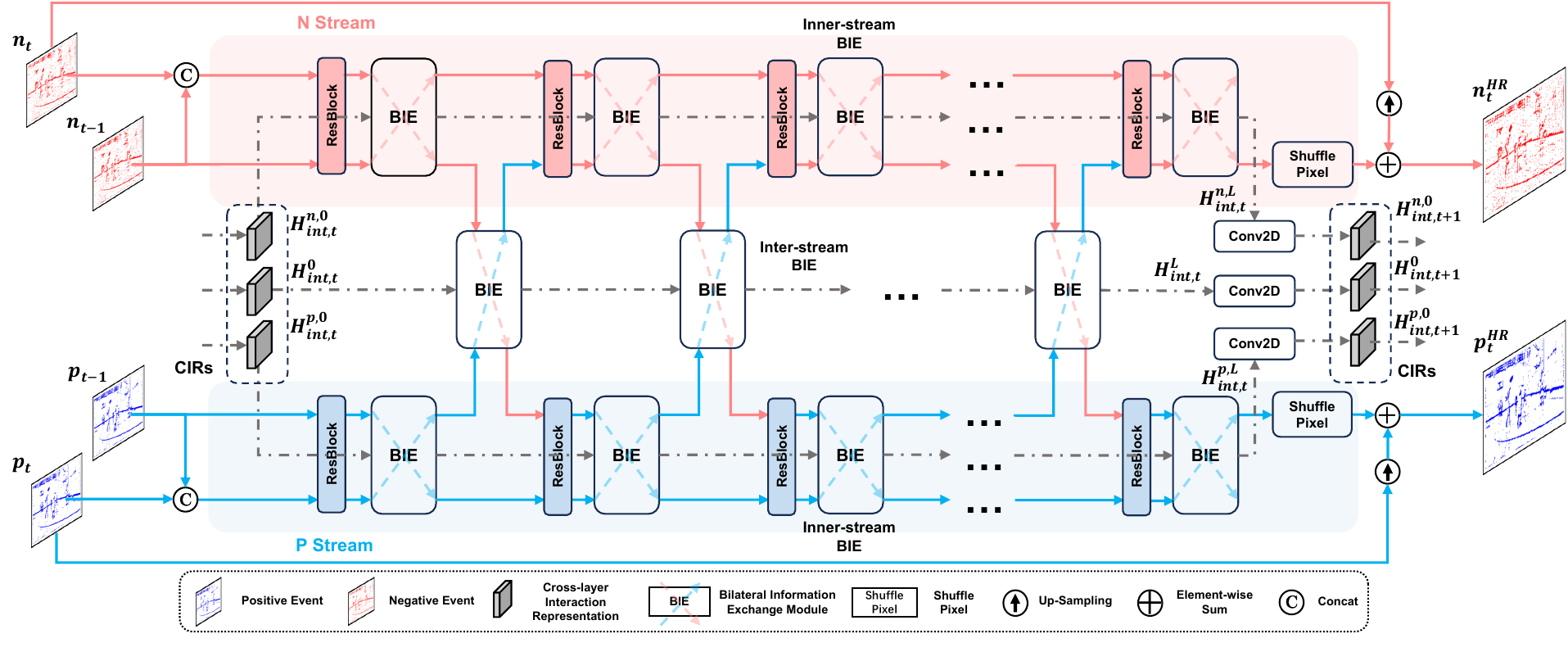}
\vspace{-4mm}
\caption{Overall framework of our \method.
Blue/red represent events with positive/negative polarity, respectively. 
\method consists of two parallel streams, each dedicated to mining the information of two events with different polarities. The inter-stream \module is applied to exchange and complement global structures between two events. Additionally, an inner-stream \module is embedded within each stream to model the spatio-temporal context of each event. Through the layer-wise introduction of inter- and inner-stream \module, \method can effectively model hierarchical spatio-temporal contextual correlations between different events, thereby enhancing the performance in ESR.
}
\vspace{-4mm}
\label{fig:arch}
\end{figure*}

To apply learning-based methods to ESR tasks, researchers proposed projecting event streams onto a 2D plane \cite{maqueda2018event, zhu2018ev} and then performing super-resolution. Although this approach compresses the temporal information of event streams, it doesn't impact tasks that require event streams to be converted into a 2D format \cite{li2023sodformer, tulyakov2021time, gehrig2019end}. Duan et al. \cite{duan2021eventzoom} were the first to convert event streams into an event stack and introduced a network based on the 3D U-Net for ESR. However, this method faces challenges in terms of memory requirements and training difficulty when handling high-factor super-resolution. Subsequently, Weng et al. \cite{weng2022boosting} presented an event stream super-resolution method based on Recurrent Neural Networks. They effectively addressed the challenges of high-factor super-resolution through the design of the temporal propagation net and spatiotemporal fusion net.
Nonetheless, these approaches overlooked the importance of decoupling event data \cite{gehrig2019end, tan2022multi} and didn't fully consider its sparsity and structural information. In light of these issues, we propose a bilateral event mining and complementary network capable of independently handling positive and negative events and facilitating mutual improvement between the two event types.

\section{Bilateral Event Mining and Complementary}
\subsection{Preliminary}
Applying learning-based methods to Event Stream Super-Resolution typically involves three steps \cite{duan2021eventzoom, weng2022boosting}. Firstly, low-resolution event streams are transformed into a 2D representation, compressing the temporal dimension of event streams. Subsequently, a super-resolution network is employed to obtain high-resolution event representations. Finally, recovering the high-resolution event stream through resampling methods. For a set of event streams, we can represent them as \(\varepsilon_n = \{\rve_k\}_{k=1}^N\), where \(N\) represents the number of events. Each \(\rve_i\in\varepsilon_n\) can be represented by a tuple \((x_i, y_i, t_i, p_i)\), where \(x_i\) and \(y_i\) represent the spatial position of the event, \(t_i\) represents the timestamp of the event, and \(p_i=\pm1\) indicates the polarity of the event. Subsequently, we partition \(\varepsilon_n\) into positive events $\{\rve_k\}^{N_p}_{k=1}$ and negative events $\{\rve_k\}^{N_n}_{k=1}$ based on their polarity \(p_i\). We transform $\{\rve_k\}^{N_p}_{k=1}$ and $\{\rve_k\}^{N_n}_{k=1}$ into event count image \cite{weng2022boosting, zhu2018ev} to describe the spatial distribution of events. Thus, we can build up two-channel event representations from \(\varepsilon_n\): positive $\rvp_k \in \R^{H\times W}$ and negative $\rvn_k \in \R^{H\times W}$. 

\subsection{Overall Pipeline}
An overview of our proposed bilateral event mining and complementary networks (\method) is depicted in \cref{fig:arch}. 
\method comprises two parallel $L$-layer streams that individually process decoupled positive and negative events. 
In each layer of \method, the bilateral information exchange (\module) module are embedded to facilitate the exchange of processed spatial information between the two streams.
We refer to this \module as the inter-stream \module.
Moreover, to fully incorporate temporal dynamics and exploit temporal contexts, we introduce another sub-stream in both P- and N-stream. 
Specifically, each sub-stream utilizes a residual block consisting of two 3 × 3 convolution layers to process spatial or temporal information.
The spatial and temporal representation are initialized from the current frame and the concatenation of two consecutive frames (the current frame and the previous one), respectively.
And the \module is introduced to layer-wisely facilitate the exchange of information between spatial and temporal contexts. We refer to this \module as the inner-stream \module. 
Both the inter- and inner-stream \module incorporate the cross-level interaction representations (CIR) that serve different roles. 
The CIR in the inner-stream \module provides the interaction between temporal and spatial information, while the CIR in the inter-stream \module restores the shared global structural information of different types of events. 
At the time step $t$, each CIR is updated through an $1\times 1$ convolutional layer and a ReLU activation function:
\begin{equation}
    \rmH_{int, t+1}^{\cdot, 0} = \text{ReLU}(\text{Conv}_{1\times1}(\rmH_{int, t}^{\cdot, L}))
\end{equation}
where $\space^{\cdot}$ denotes the type of CIR. $\rmH_{int, t}^{p,L}$, $\rmH_{int, t}^{n,L}$ and $\rmH_{int, t}^{L}$ are CIRs from the last layer of inner-stream \module in P-, N-stream and the last layer of inter-stream \module, respectively. 
And at each time step, both CIRs of inner- and inter-stream \module are initialized by the updated CIR from previous time step. 
Particularly, at time step $0$, CIRs are initialized as empty representations. 
This design enables hierarchical information propagation within \method, enhancing its ability to capture spatial-temporal contextual information. 
After extracting deep-level features, we utilized pixel shuffle \cite{shi2016real} to transform the feature information into high-resolution event count image \({E}_{t}^{SR}\). Finally, high-resolution event streams can be obtained by resampling.

\begin{figure}[htp]
\centering
\includegraphics[width=0.95\linewidth]{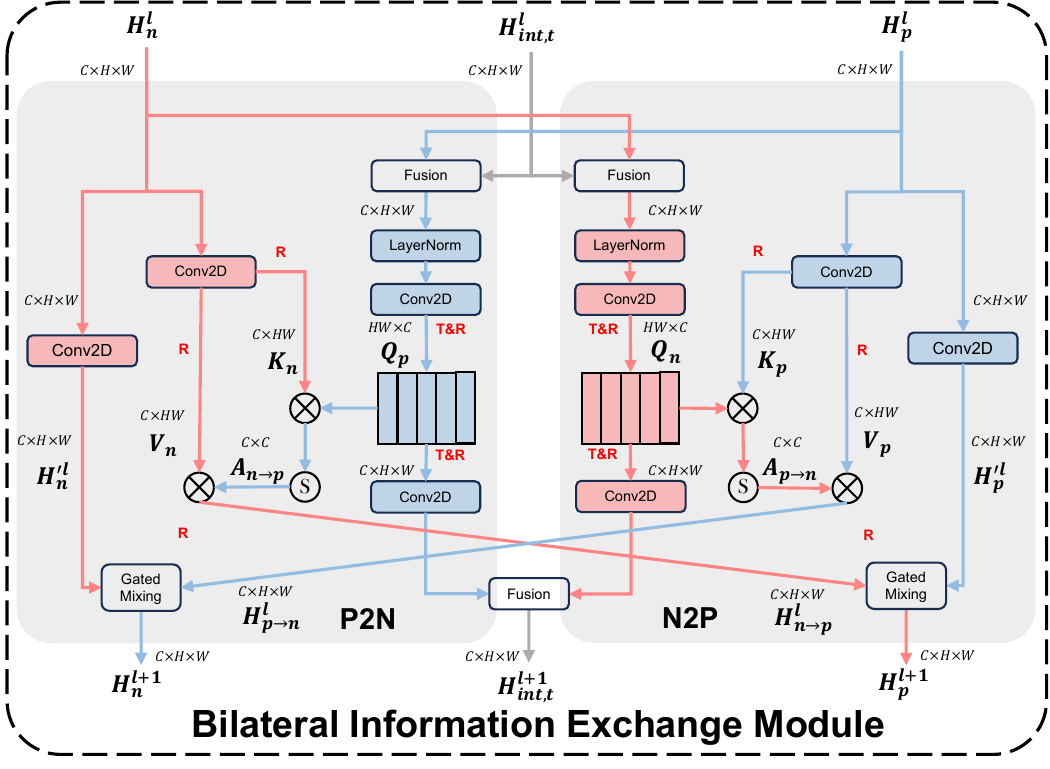}
\caption{The architecture of the proposed bilateral information exchange (\module) module. \textbf{[Best viewed with zoom-in.]}}
\label{fig:rsm}
\vspace{-2mm}
\end{figure}
\subsection{Bilateral Information Exchange Module}
Since the positive and negative events exhibit high correlation in corresponding spatial locations and their neighboring regions, the global structures in two types of events can mutually benefit each other by means of providing complementary information.
To facilitate the selective integration of global structural information between two types of events and mitigate the potential misleading effects of noises, we propose a bilateral information exchange (\module) module, as shown in \cref{fig:rsm}.
For the event at time step $t$, we denote the representations of positive and negative events in $l$-th layer of \method as $\rmH_p^l \in \mathbb{R}^{C\times H \times W}$ and $\rmH_n^l \in \mathbb{R}^{C\times H \times W}$, respectively. 
And a cross-level interaction representation (CIR) $\rmH_{int, t}^l \in \mathbb{R}^{C\times H\times W}$ is introduced to incorporate the hierarchical contexts into the information exchanging process.
In this subsection, we describe the details of the information propagation from the positive to the negative event, and the similar process is applied in the propagation from the negative event to the positive one.

Firstly, the CIR $\rmH_{int, t}^l$ is updated through the LayerNorm and an $1\times 1$ convolution, and then the updated CIR is fused with the $\rmH_n^l$ through another $1\times 1$ convolutional layer, obtained the reshaped query $\rmQ_n^l \in \mathbb{R}^{M\times HW}$. Here $M$ represent the number of global structures we assumed. Then, $\rmH_{p}^l$ is clustered into $\rmV_p^l \in \mathbb{R}^{M\times HW}$ and $\rmK_p^l \in \mathbb{R}^{M\times HW}$ through the stacked convolutional layers and reshape operation.
By calculating the correlation of $\rmQ_n^l$ and $\rmK_p^l$ and applying the Softmax function along the horizontal dimension, we can obtain the attention scores matrix $\rmA_{p\shortrightarrow n} \in \mathbb{R}^{M\times M}$ which reflects the correlations between $M$ semantics in $\rmQ_n^l$ and $\rmK_p^l$. 
Following Scaled Dot-Product Attention \cite{vaswani2017attention}, the scale coefficient $\sqrt{HW}$ is applied in the calculation process.
The attention scores are then utilized to aggregate the global structural representations in $\rmV_p^l$, obtaining the $\rmH_{p\shortrightarrow n}^{'l} \in \mathbb{R}^{M\times HW}$. $\rmH_{p\shortrightarrow n}^{'l}$ is further reshaped and projected into $\rmH_{p\shortrightarrow n}^{l} \in \mathbb{R}^{C\times H\times W}$ through an $1\times 1$ convolutional layer. 
Besides, a residual branch consists of stacked convolutional layers is introduced to process the original information of $\rmH_n^l$ for the local structural details, obtained the $\rmH_n^{'l}$. The process is formulated as:
\begin{equation}
    \rmH_{p\shortrightarrow n}^l = \text{Conv}_{1\times 1}(\text{Softmax}(\rmQ_n^l {(\rmK^l_p)}^{T} / \sqrt{HW}) \rmV_p^l)
\end{equation}
Finally, the $\rmH_n^{l}$ is updated by fusing $\rmH_{p\shortrightarrow n}^{'l}$ and $\rmH_n^{'l}$ through a gated mixing module:
\begin{equation}
    \rmH_n^{l+1} = \rmZ_n^l \odot \rmH_n^{'l} + (1 - \rmZ_n^l) \odot \rmH_{p\shortrightarrow n}^{l}
\end{equation}
\vspace{-3mm}
\begin{equation}
    \rmZ_n^l = \sigma(\rmW_{n1} \rmH_n^{'l} + \rmW_{n2} \rmH_{p\shortrightarrow n}^{l})
\end{equation}
where $\sigma$ denotes the Sigmoid function whose range lies in $[0,1]$, $\odot$ denotes element-wise production and $\rmZ_n^l \in \mathbb{R}^{}$.
The similar procedure of information propagation from the negative to the positive event can be written directly:
\begin{equation}
    \rmH_{n\shortrightarrow p}^l = \text{Conv}_{1\times 1}(\text{Softmax}(\rmQ_p^l {(\rmK^l_n)}^{T} / \sqrt{HW}) \rmV_n^l)
\end{equation}
\vspace{-3mm}
\begin{equation}
    \rmH_p^{l+1} = \rmZ_p^l \odot \rmH_p^{'l} + (1 - \rmZ_p^l) \odot \rmH_{n\shortrightarrow p}^{l}
\end{equation}
\vspace{-3mm}
\begin{equation}
    \rmZ_p^l = \sigma(\rmW_{p1} \rmH_p^{'l} + \rmW_{p2} \rmH_{n\shortrightarrow p}^{l})
\end{equation}
And the CIR $\rmH_{int,t}^l$ is updated by feeding the concatenation of the reshaped $\rmQ_p^l$ and $\rmQ_n^l$ into the convolutional layer:
\begin{equation}
    \rmH_{int,t}^{l+1} = \text{Conv}(\llbracket \rmQ_p^{l,r}, \rmQ_n^{l,r} \rrbracket) + \rmH_{int}^{l}
\end{equation}
where $\llbracket \cdot \rrbracket$ denotes concatenation along channel, $r$ denotes the query matrix is reshaped from $\mathbb{R}^{C\times HW}$ into $\mathbb{R}^{C\times H\times W}$.
And the pseudo-code of \module is provided in \textbf{\cref{pseudo-code}}.

\subsection{Training Objectives}
In the training process, to ensure the continuity of the event stream, following the approach of \cite{weng2022boosting}, we divide the event stream into segments of length \(T (T=9)\) and calculate the mean square error for each segment using a sliding window:
\begin{equation}\label{eq2}
\mathcal{L} = ~{\sum_{t = 1}^{T}{MSE\left( {E}_{t}^{SR},~{E}_{t}^{HR} \right)}}
\end{equation}
where \({E}_{t}^{SR}\) represents the event count image obtained through \method, \({E}_{t}^{HR}\) represents the ground truth event count image, and \(MSE\) is the mean square error function. 
\section{Experiments}
\subsection{Datasets and Training Settings}
In this work, we validate our approach using both real and synthetic data. The real event dataset, including multi-scale LR-HR pairs, is limited due to the difficulty of aligning both temporal and spatial information simultaneously. EventNFS \cite{duan2021eventzoom} is the first real dataset involving LR-HR pairs, captured by a DAVIS346 monochromatic camera. However, due to device resolution constraints, the minimum resolution is \(55\times31\), and only \(2\times\) and \(4\times\) data pairs are available. To obtain high-quality LR-HR pairs with higher multiples, following \cite{weng2022boosting}, we use an event simulator \cite{lin2022dvs} on the NFS \cite{kiani2017need} dataset and RGB-DAVIS dataset \cite{wang2020joint} to transform them into event streams, resulting in two new synthetic datasets, NFS-syn and RGB-syn. 

To ensure a fair comparison, we maintain consistency with the training settings from \cite{weng2022boosting}. We set the batch size to 2, the initial learning rate to 0.001, a decay factor of 0.95, and decay every 4000 iterations. All models were trained for 100,000 iterations, and the entire experimentation was conducted on a Tesla V100 GPU.
For more details on dataset processing, ablation studies and experimental results, please refer to the \textbf{\cref{app-config}}.

\subsection{Comparison with State-of-the-Art Models}
In the field of ESR, our primary comparisons are made with two previous learning-based methods, EventZoom \cite{duan2021eventzoom} and RecEvSR \cite{weng2022boosting}. Comparing with other ESR methods is challenging \cite{li2021event, li2019super, wang2020joint}, as they either require real frames as assistance or may fail in complex scenarios, making a fair comparison difficult. EventZoom \cite{duan2021eventzoom}, the first learning-based method in ESR, faced training difficulties with its 3D-Unet architecture for large-factor SR. Following prior practices \cite{weng2022boosting}, we ran EventZoom-\(2\times\) multiple times to obtain results for large-factor SR. RecEvSR \cite{weng2022boosting} overcomes the challenge of large-scale SR using recurrent neural networks, and we retrained it using the provided code. Additionally, we included two representative methods for video and image super-resolution for comparison: bicubic and SRFBN \cite{li2019feedback} for image super-resolution and lightweight models RLSP \cite{fuoli2019efficient} and RSTT \cite{geng2022rstt} for video super-resolution. 
To fully investigate the effectiveness of the proposed paradigm that process two events independently and mutually refine each other, we introduced \method-plain which is obtained by removing the inner-stream \module in \method. 
We utilize the RMSE, model parameters and FLOPs to evaluate each method from three perspectives: the performance, model and computational complexity.

\begin{figure*}[ht]
  \centering
  \includegraphics[width=0.95\textwidth]{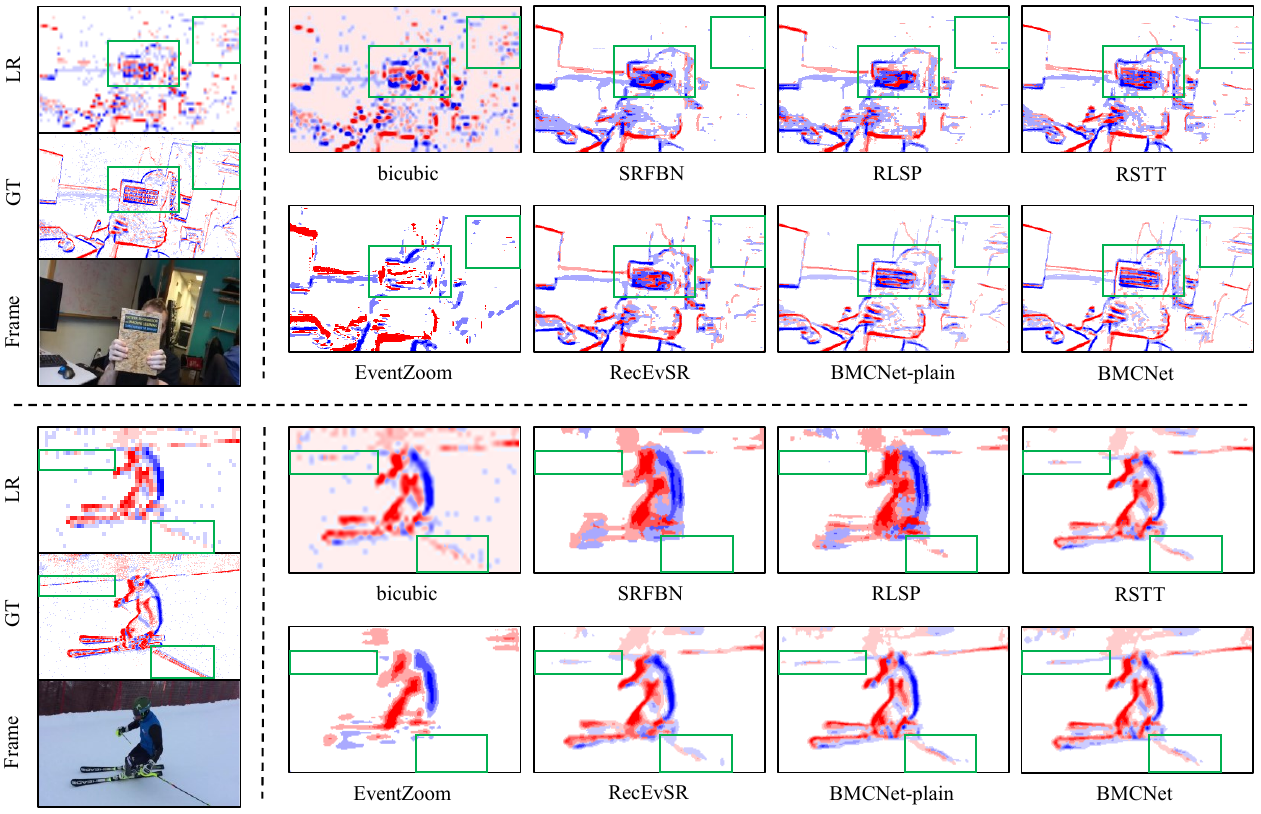}
  \caption{Qualitative analysis comparison on synthetic and real datasets. The upper and lower sections represent the \(4\times\) super-resolution results for NFS-syn and EventNFS, respectively. "GT" denotes the \(4\times\) ground-truth. Our \method-plain and \method demonstrates superior detail recovery and clearer edges on both datasets. \textbf{[Best viewed with zoom-in.]}}
  \label{figure2}
\end{figure*}

\noindent\textbf{Qualitative Analysis Results.} \cref{figure2} presents the \(4\times\) SR results of various methods on both synthetic and real datasets (please refer to \textbf{\cref{app-more-quali-results}} for more results). 
It is evident that bicubic interpolation struggles to effectively contribute to ESR. EventZoom exhibits numerous issues with detail loss, potentially arising from error accumulation in multiple runs of EventZoom-\(2\times\). While SRFBN and RLSP can restore the overall structures of event count images, they suffer from significant detail loss and blurred object edges. In comparison, RSTT and RecEvSR perform reasonably in recovering event count image. However, they still fall short in detail repair. Contrasted with these methods, our \method-plain and \method excel in mutually complementing and repairing details by leveraging overall structural information from two events, resulting in richer details and clearer edges.

\noindent\textbf{Quantitative Analysis Results.} As shown in \cref{tab:main_tab}, both \method-plain and \method have achieved state-of-the-art (SOTA) performance across all datasets. Compared to RecEvSR, the previous SOTA method in ESR, our \method-plain exhibits an average improvement of 7.5\% in super-resolution across all datasets, accompanied by a 40\% reduction in parameters. Furthermore, in comparison to the leading Transformer-based video super-resolution method, RSTT, our \method-plain achieves an average improvement of 3.5\%, with a reduction in parameters of 73\%. Meanwhile, \method, which includes both inter- and inner-stream \module, demonstrates superior performance. \method boasts an average improvement of 11\% over RecEvSR, but with a parameters increase of 50\%. In contrast, \method achieves an average improvement of 7\% over RSTT, with a reduction in parameters of 31\%. Additionally, FLOPs values indicate that our \method-plain exhibits the lowest average complexity, while \method maintains competitive one. These findings collectively underscore the efficiency of our methods.

\begin{table*}
\centering
\begin{adjustbox}{width=0.9\textwidth}
    \renewcommand{\arraystretch}{1.2}
\begin{tabular}{lccclcccccclccccccc}
\toprule
\multirow{2}{*}{Methods} &
  \multicolumn{3}{c}{NFS-syn} &
   &
  \multicolumn{2}{c}{RGB-syn} &
   &
  \multicolumn{2}{c}{EventNFS-real} &
   &
  \multicolumn{3}{c}{\# Param (M)} &
   &
  \multicolumn{3}{c}{\# FLOPs (G)}
  \\ \cmidrule{2-4} \cmidrule{6-7} \cmidrule{9-10} \cmidrule{12-14} \cmidrule{16-18}
  & \(2\times\)    & \(4\times\)    & \(8\times\)    &  & \(2\times\)    & \(4\times\)    &  & \(2\times\)    & \(4\times\) &  & \(2\times\)    & \(4\times\)       & \(8\times\) &  & \(2\times\)    & \(4\times\)       & \(8\times\) \\ 
  \midrule

bicubic   & 0.785 & 0.729 & 0.738 &  & 0.346 & 0.378 &  & 0.872 & 0.948 & & - & - & -  & & - & - & - \\
\midrule
SRFBN-esr \cite{li2019feedback}   & 0.641 & 0.628 & 0.628 &  & 0.301 & 0.300 &  & 0.644 & 0.738  & & 2.1 & 3.6 & 7.9 & & 39.5 & 116.7 & 984.1 \\
RLSP-esr \cite{fuoli2019efficient}  & 0.642 & 0.624 & 0.621 & & 0.298 & 0.294 & & 0.623 & 0.705 & & \underline{1.2} & \underline{1.2} & \underline{1.5} & & 23.1 & 24.9 & \underline{32.1}    \\
RSTT-esr \cite{geng2022rstt}      & 0.624 & 0.605 & 0.604 &  & 0.298 & 0.292 &  & 0.557 & 0.632 & & 3.8 & 4.1 & 4.3 & & 22.3 & 43.3 & 61.4 \\
\midrule
EventZoom \cite{duan2021eventzoom} & 0.898 & 1.024 & 1.113 &  & 0.479 & 0.975 &  & 0.882 & 1.117 & & 11.5 & 11.5 & 11.5 & & 65.3 & 81.0 & 220.3 \\
RecEvSR \cite{weng2022boosting}   & 0.656 & 0.607 & \underline{0.576} &  & 0.319 & 0.296 & & 0.613 & 0.670 & & 1.8 & 1.8 & 1.8 & & \textbf{2.8} & \underline{10.7} & 42.1 \\
\midrule
\method-plain (Ours) & \underline{0.592} & \underline{0.577} & 0.579 &  & \underline{0.287} & \underline{0.285} &  & \underline{0.541} & \textbf{0.619} & & \textbf{0.9} & \textbf{1.0} & \textbf{1.4} & & \underline{7.96} & \textbf{8.16} & \textbf{8.96} \\ 

\method (Ours) & \textbf{0.564} & \textbf{0.552} & \textbf{0.553} & & \textbf{0.276} & \textbf{0.274} & & \textbf{0.527} & \underline{0.625} & \textbf{} & 2.6 & 2.7 & 3.1  & & 35.35 & 35.65 & 36.84 \\

\bottomrule
\end{tabular}
\renewcommand{\arraystretch}{1}
\end{adjustbox}
\caption{Quantitative analysis comparison on real and synthetic datasets, and RMSE, model parameters and FLOPs are reported. \method denotes the network additionally equipped with both the inter- and inner-stream \module for spatial and temporal contexts modeling, and \method-plain denotes only the inter-\module is equipped as described in \cref{BIE}. The FLOPs is calculated on the LR events of NFS-syn dataset with resolutions of 80$\times$45. Top 2 results are highlighted with \textbf{bold text} and \underline{underlined text}, respectively.}
\vspace{-3mm}
\label{tab:main_tab}
\end{table*}

\subsection{Model Analysis}
\label{BIE}
\noindent\textbf{The Validation of Main Components in \method.}
We conducted a series of ablation experiments on the NFS-syn datasets to investigate the impact of the main components in \method. 
The experimental results are presented in \cref{tab:abla}, 
\textbf{Decoup.} refers to the model processing positive and negative events decoupled from the event stream separately, 
\textbf{Rec.} signifies the introduction of a recurrent hidden state into the model for cross-frame modeling, 
\textbf{Inter-\module} indicates the layer-wise application of inter-stream \module for global structures exchange between two types of events, 
and \textbf{Inner-\module} denotes the inner-stream \module embedded in the model for spatial and temporal contexts modeling.

By comparing Exp0 and Exp1, we observed that the inclusion of a recurrent hidden state improves performance on both synthetic and real event datasets. This improvement can be attributed to the capture of cross-frame correlations and the utilization of useful cues from previous frames for global structure understanding. 
Additionally, the comparisons between Exp1-4 demonstrate that simply processing decoupled positive and negative events individually can not model the correlations between two events, resulting in bad performance.
Compared to directly process positive and negative events in a mixed paradigm, separately process decoupled events and exchange global structures through \module can effectively capture inner- and inter-event information, enhancing the performance in super-resolutions. 
More ablation studies about comparison between mixed and decoupled paradigm please refer to subsections below.
Furthermore, the comparison between Exp4 and Exp5 demonstrates that simultaneously modeling the temporal-spatial information correlations and leveraging the complementary nature of the two types of events can fully exploit spatial and temporal contextual information in the event streams. This approach significantly promotes the final performance of the model in event super-resolution.

\begin{table}[ht]
\centering
\begin{adjustbox}{width=0.9\linewidth}
    \renewcommand{\arraystretch}{1.2}
\begin{tabular}{l|cccc|ccc}
\toprule
Exps & Decoup. & Rec. & \thead{Inter-\\\module}  & \thead{Inner-\\\module} & NFS-syn & EventNFS-real & \# Params.\\ 
\midrule
Exp0 & \textcolor{lightgray}{\ding{56}} & \textcolor{lightgray}{\ding{56}} & \textcolor{lightgray}{\ding{56}} & \textcolor{lightgray}{\ding{56}} & 0.605 & 0.655 & \textbf{0.38} M \\
Exp1 & \textcolor{lightgray}{\ding{56}} & \ding{52} & \textcolor{lightgray}{\ding{56}} & \textcolor{lightgray}{\ding{56}} & 0.599 & 0.648 & \underline{0.68} M\\
Exp2 & \ding{52} & \ding{52} & \textcolor{lightgray}{\ding{56}} & \textcolor{lightgray}{\ding{56}} & 0.637 & 0.768 & 0.97 M \\
Exp3 & \ding{52} & \textcolor{lightgray}{\ding{56}} & \ding{52} & \textcolor{lightgray}{\ding{56}} & 0.580 & 0.641 & 0.94 M \\
Exp4 & \ding{52} & \ding{52} & \ding{52} & \textcolor{lightgray}{\ding{56}} & \underline{0.577} & \textbf{0.619} & 1.00 M \\
Exp5 & \ding{52} & \ding{52} & \ding{52} & \ding{52} & \textbf{0.552} & \underline{0.625} & 2.72 M \\
\bottomrule
\end{tabular}
\renewcommand{\arraystretch}{1}
    \end{adjustbox}
\caption{Quantitative analysis of branch ablation experiments for \(4\times\) SR on RMSE metrics.}
\vspace{-3mm}
\label{tab:abla}
\end{table}

\begin{figure}[ht]
  \centering
  \includegraphics[width=0.97\linewidth]{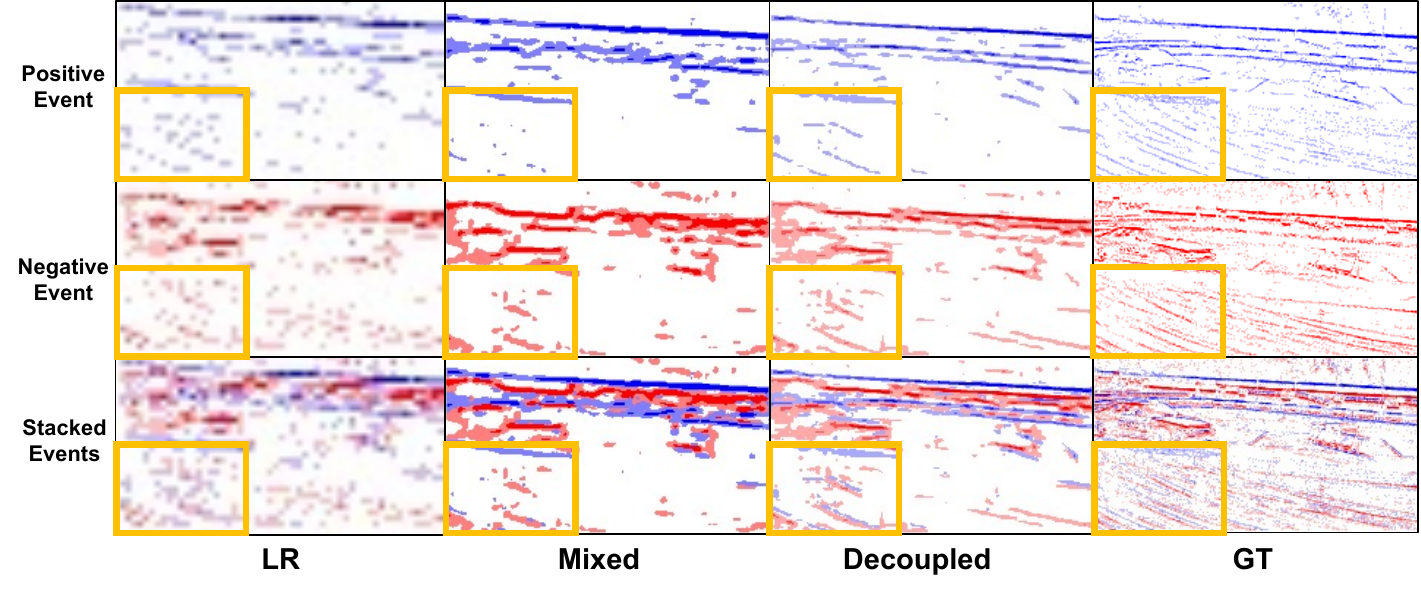}
  \vspace{-3mm}
  \caption{Qualitative comparison on the real dataset EventNFS-real between the mixed and decoupled paradigm.}
  \label{fig:abla-decoup}
  \vspace{-2mm}
\end{figure}

\begin{figure}[t]
  \centering
  \includegraphics[width=\linewidth]{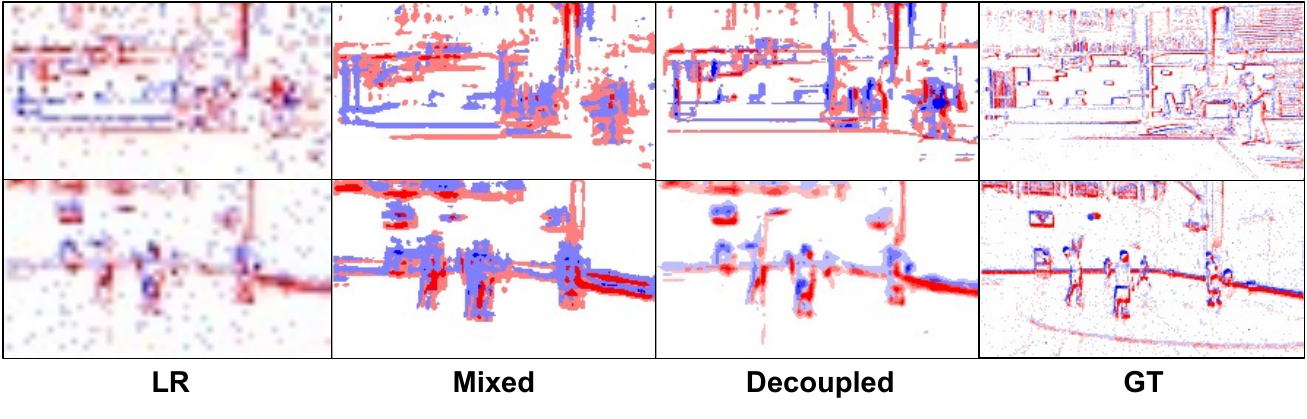}
  \vspace{-5mm}
  \caption{Artifacts of overlapped events appear when two events are processed in a mixed paradigm.}
  \label{fig:abla-decoup2}
  \vspace{-3mm}
\end{figure}

\noindent\textbf{Benefit from Decoupling Positive and Negative Events.}
The core motivation behind our \method is to fully exploit the correlations between positive and negative events, allowing them to benefit each other by providing the complementary information. 
Through the decoupled paradigm, the model has capability to capture unique characteristics of each event and enable effective interactions between them. 
And overall performance can be enhanced through leveraging the correlation and complementary nature of two events.
To verify this, we conduct experiments that process positive and negative events in both the mixed and decoupled paradigms, respectively. The comparison results are presented in \cref{tab:abla} and \cref{fig:abla-decoup}.
Moreover, we observed that simultaneously processing positive and negative events in the mixed paradigms always results in overlapped artifacts. The reason for this is that simultaneous processing of positive and negative events makes the model confused in distinguishing between the two types of events. This phenomenon is resolved when we process the positive and negative events in a decoupled paradigm, as shown in \cref{fig:abla-decoup2}.

\begin{figure*}[ht]
  \centering
  \includegraphics[width=0.95\textwidth]{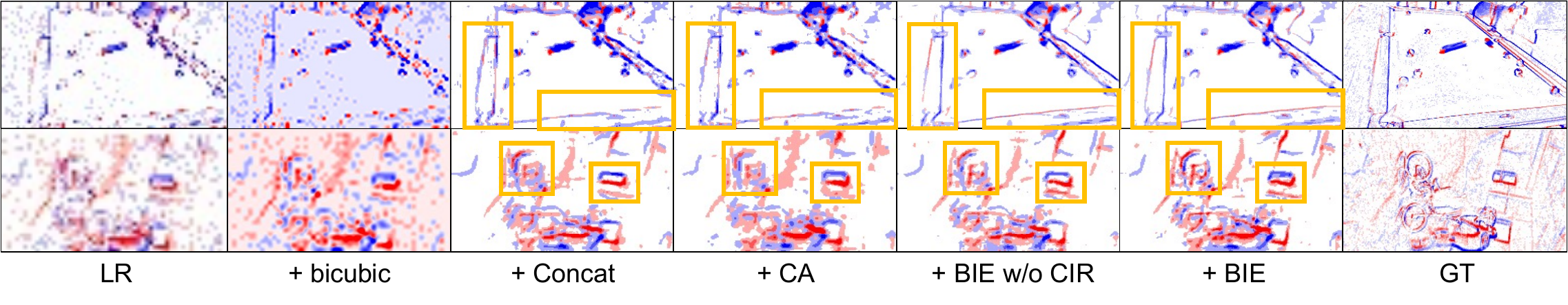}
  \caption{Qualitative analysis comparison between bicubic, Concat, CA, \module with and without CIR on synthetic and real datasets.}
  \label{fig:abla-nrsm}
\end{figure*}

\noindent\textbf{Effectiveness of Bilateral Information Exchange Module.}
In this section, we use the \method-plain (equipped with inter-stream \module only) as baseline to conduct ablation studies on NFS-syn and EventNFS-real datasets for investigating the \module we proposed. 
The \module serves as the core operation in our \method, allowing for the effective exchange of global structural information between positive and negative events without being misled by noise. 
And the introduction of the cross-level interaction representation (CIR) enables the model to exchange information across different levels.
To verify the effectiveness of \module, we replace \module in \method-plain with three different type operations: (1) concatenation, (2) cross attention (CA) operation, (3) the \module without CIR.
The quantitative comparisons are presented in \cref{tab:abla-nrsm}, which are demonstrated the effectiveness the \module and CIR on improving the overall performance.
Additionally, the qualitative comparisons are shown in \cref{fig:abla-nrsm}.
The model exchanges information between two events through the concatenation cannot effectively utilize the complementary information, resulting in incomplete structures.
The model equipped with CA is easily misled by noises which results in distort structures and artifacts.
On the contrary, our \module has capability to alleviate the impact of invalid information and reconstruct clear edges by treating each channel as a structural representation and exchanging it along the channel dimension. 

\begin{table}[ht]
\centering
\begin{adjustbox}{width=0.93\linewidth}
    \renewcommand{\arraystretch}{1.2}
\begin{tabular}{l|cccc}
\toprule
Metrics & NFS-syn & EventNFS-real & \# Params & Complexity\\ 
\midrule
Concat. & 0.606 & 0.664 & 1.40 M & $\mathcal{O}(C^2)$ \\
CA & 0.610 & 0.643 & 1.37 M & $\mathcal{O}(CH^2W^2)$ \\
BIE w/o CIR & \underline{0.580} & \underline{0.641} & \textbf{0.94} M & $\mathcal{O}(C^2HW)$ \\
\module & \textbf{0.577} & \textbf{0.619} & \underline{1.00} M & $\mathcal{O}(C^2HW)$ \\
\bottomrule
\end{tabular}
\renewcommand{\arraystretch}{1}
    \end{adjustbox}
\caption{Effectiveness of \module and CIR. The experiments are conducted for \(4\times\) SR and evaluated on RMSE metrics.}
\vspace{-3mm}
\label{tab:abla-nrsm}
\end{table}

\subsection{Enhancing Downstream Applications}
\begin{table}[t]
\centering
\begin{adjustbox}{width=0.99\linewidth}
    \renewcommand{\arraystretch}{1.2}
\begin{tabular}{lllllllll}
\toprule
          & \multicolumn{8}{c}{Object Recognition}              \\
\multirow{2}{*}{Methods} & \multicolumn{2}{c}{2x}          &           & \multicolumn{2}{c}{4x}          &           & \multicolumn{2}{c}{8x}          \\ \cmidrule{2-3} \cmidrule{5-6} \cmidrule{8-9} 
          & ACC \(\uparrow\)   & AUC \(\uparrow\)   &  & ACC \(\uparrow\)   & AUC \(\uparrow\)  &  & ACC \(\uparrow\)  & AUC \(\uparrow\)  \\ \midrule
bicubic   & 56.67 & 57.43 &  & 56.01 & 56.89 &  & 49.95 & 50.77 \\
SRFBN     & 61.12 & 61.94 &  & 60.89 & 61.03 &  & 50.02 & 50.86 \\
RLSP      & 61.33 & 62.16 &  & 61.19 & 61.25 &  & 50.26 & 50.97 \\
RSTT      & 63.51 & 63.96 &  & 63.02 & 64.29 &  & 52.97 & 54.07 \\
EventZoom & 54.68 & 56.03 &  & 49.56 & 50.45 &  & 47.96 & 48.74 \\
RecEvSR   & 62.91 & 63.47 &  & 62.37 & 63.07 &  & 53.57 & 54.48 \\
BMCNet-plain    & \underline{66.95} & \underline{67.73} &  & \underline{67.31} & \underline{67.63} &  & \underline{54.31} & \underline{54.93} \\
BMCNet & \textbf{68.58} & \textbf{69.33} &  & \textbf{68.95} & \textbf{69.23} &  & \textbf{57.97} & \textbf{58.86} \\ \midrule
GT                       & 85.16 & 84.99 &  & 93.44 & 93.52 &  & 94.96 & 94.81 \\ \bottomrule
\end{tabular}
\renewcommand{\arraystretch}{1}
\end{adjustbox}
\begin{adjustbox}{width=0.99\linewidth}
    \renewcommand{\arraystretch}{1.2}
\begin{tabular}{lllllllll}
\toprule
\multirow{2}{*}{Methods} & \multicolumn{8}{c}{Video Reconstruction}            \\ \cmidrule{2-9} 
                         & SSIM \(\uparrow\)  & LPIPS \(\downarrow\) &  & SSIM \(\uparrow\)  & LPIPS \(\downarrow\) &  & SSIM\(\uparrow\)  & LPIPS \(\downarrow\) \\ \midrule
bicubic                  & 0.568 & 0.395 &  & 0.609 & 0.522 &  & 0.598 & 0.545 \\
SRFBN                    & 0.608 & 0.389 &  & 0.618 & 0.455 &  & 0.612 & 0.489 \\
RLSP                     & 0.603 & 0.384 &  & 0.620 & 0.439 &  & 0.617 & 0.476 \\
RSTT                     & 0.627 & 0.359 &  & 0.639 & 0.424 &  & 0.622 & 0.472 \\
EventZoom                & 0.542 & 0.429 &  & 0.575 & 0.488 &  & 0.574 & 0.542 \\
RecEvSR                  & 0.611 & 0.371 &  & 0.637 & 0.426 &  & 0.630 & \underline{0.466} \\
BMCNet-plain                   & \underline{0.638} & \underline{0.349} &  & \underline{0.651} & \underline{0.418} &  & \underline{0.632} & 0.468 \\
BMCNet & \textbf{0.646} & \textbf{0.343} &  & \textbf{0.661} & \textbf{0.414} &  & \textbf{0.652} & \textbf{0.452} \\ \bottomrule
\end{tabular}
\renewcommand{\arraystretch}{1}
\end{adjustbox}
\caption{Quantitative analysis results on downstream tasks of object recognition and video reconstruction. For object recognition, the evaluation is conducted on the NCars dataset \cite{sironi2018hats}, where AUC and ACC represent accuracy and area under the curve, respectively. GT denotes the reference result obtained by directly recognizing downsampled event streams. Video reconstruction is performed on the NFS-syn dataset. \textbf{Bold} and \underline{underline} indicate the best and the second-best performance.}
\vspace{-3mm}
\label{tab:da_1}
\end{table}

\textbf{Object Recognition.} We investigated the performance of bicubic, SRFBN \cite{li2019feedback}, RLSP \cite{fuoli2019efficient}, RSTT \cite{geng2022rstt}, EventZoom \cite{duan2021eventzoom}, RecEvSR \cite{weng2022boosting}, and our \method-plain and \method in the context of object recognition applications. Utilizing the classifier proposed by Gehrig et al. \cite{gehrig2019end}, we conducted object recognition on the NCars dataset \cite{sironi2018hats}. Initially, the NCars dataset underwent an \(8\times\) downsampling, followed by applying each model trained on the NFS-syn dataset to perform \(2(4,8)\times\) super-resolution. The super-resolved data was then used for recognition, and the results were compared. We utilized accuracy (ACC) and area under the curve (AUC) as the metrics for comparison. The GT represents the classification results obtained by directly downsampling the event stream to the same resolution, indicating the upper limit performance. \cref{tab:da_1} presents the quantitative analysis results for object recognition, demonstrating that our \method-plain and \method outperform other approaches significantly across \(2(4,8)\times\) super-resolution scales, validating the robust detail recovery capability of our methods.

\noindent\textbf{Video Reconstruction.} Video reconstruction is a crucial task in event-based vision \cite{rebecq2019high, stoffregen2020reducing, weng2021event}. We also compared all methods in this task. Initially, we applied each model to \(2(4,8)\times\) super-resolution on the NFS-syn dataset, which was downsampled by a factor of 16. Then, we utilized the E2VID \cite{rebecq2019high} for reconstructing images based on the super-resolved event stream. Finally, we evaluated the reconstructed images using the structural similarity (SSIM) \cite{wang2004image} and the perceptual similarity (LPIPS) \cite{zhang2018unreasonable} metrics. Please refer to \textbf{\cref{app-more-downstream-results}} for reconstructed images. \cref{tab:da_1} presents quantitative comparisons for reconstructed images, indicating that our \method-plain and \method outperform other approaches significantly at different scales.
\section{Conclusion}
In this paper, we fully consider the inherent characteristics of the event data and propose a novel bilateral event mining and complementary network (\method) for the ESR task. By simultaneously modeling the distinct data distribution of positive and negative events which are decoupled from the event stream, and capturing the complementary information to mutually refine each other, \method can effectively reconstruct clear structures of each type of event.
Extensive empirical studies and analysis experiments conducted on two synthetic and one real datasets demonstrate the effectiveness and superiority of \method. Moreover, \method are evaluated on two downstream tasks, achieving outstanding results and further highlighting its superiority over other approaches.
\section*{Acknowledgments}
This work was partly supported by the National Natural Science Foundation of China (No.62171251 \& 62311530100), the Special Foundations for the Development of Strategic Emerging Industries of Shenzhen (No.JSGG20211108092812020) and the Major Key Research Project of PCL (No.PCL2023A08).

{
    \small
    \bibliographystyle{ieeenat_fullname}
    \bibliography{main}

\begin{thebibliography}{45}
\providecommand{\natexlab}[1]{#1}
\providecommand{\url}[1]{\texttt{#1}}
\expandafter\ifx\csname urlstyle\endcsname\relax
  \providecommand{\doi}[1]{doi: #1}\else
  \providecommand{\doi}{doi: \begingroup \urlstyle{rm}\Url}\fi

\bibitem[Bardow et~al.(2016)Bardow, Davison, and Leutenegger]{bardow2016simultaneous}
Patrick Bardow, Andrew~J Davison, and Stefan Leutenegger.
\newblock Simultaneous optical flow and intensity estimation from an event camera.
\newblock In \emph{Proceedings of the IEEE conference on computer vision and pattern recognition}, pages 884--892, 2016.

\bibitem[Brandli et~al.(2014)Brandli, Berner, Yang, Liu, and Delbruck]{brandli2014240}
Christian Brandli, Raphael Berner, Minhao Yang, Shih-Chii Liu, and Tobi Delbruck.
\newblock A 240$\times$ 180 130 db 3 $\mu$s latency global shutter spatiotemporal vision sensor.
\newblock \emph{IEEE Journal of Solid-State Circuits}, 49\penalty0 (10):\penalty0 2333--2341, 2014.

\bibitem[Cannici et~al.(2020)Cannici, Ciccone, Romanoni, and Matteucci]{cannici2020differentiable}
Marco Cannici, Marco Ciccone, Andrea Romanoni, and Matteo Matteucci.
\newblock A differentiable recurrent surface for asynchronous event-based data.
\newblock In \emph{Computer Vision--ECCV 2020: 16th European Conference, Glasgow, UK, August 23--28, 2020, Proceedings, Part XX 16}, pages 136--152. Springer, 2020.

\bibitem[Chan et~al.(2021)Chan, Wang, Yu, Dong, and Loy]{chan2021basicvsr}
Kelvin~CK Chan, Xintao Wang, Ke Yu, Chao Dong, and Chen~Change Loy.
\newblock Basicvsr: The search for essential components in video super-resolution and beyond.
\newblock In \emph{Proceedings of the IEEE/CVF conference on computer vision and pattern recognition}, pages 4947--4956, 2021.

\bibitem[Chan et~al.(2022)Chan, Zhou, Xu, and Loy]{chan2022basicvsr++}
Kelvin~CK Chan, Shangchen Zhou, Xiangyu Xu, and Chen~Change Loy.
\newblock Basicvsr++: Improving video super-resolution with enhanced propagation and alignment.
\newblock In \emph{Proceedings of the IEEE/CVF conference on computer vision and pattern recognition}, pages 5972--5981, 2022.

\bibitem[Chen et~al.(2023)Chen, Wang, Zhou, Qiao, and Dong]{chen2023activating}
Xiangyu Chen, Xintao Wang, Jiantao Zhou, Yu Qiao, and Chao Dong.
\newblock Activating more pixels in image super-resolution transformer.
\newblock In \emph{Proceedings of the IEEE/CVF Conference on Computer Vision and Pattern Recognition}, pages 22367--22377, 2023.

\bibitem[Duan et~al.(2021)Duan, Wang, Zhou, Ma, and Shi]{duan2021eventzoom}
Peiqi Duan, Zihao~W Wang, Xinyu Zhou, Yi Ma, and Boxin Shi.
\newblock Eventzoom: Learning to denoise and super resolve neuromorphic events.
\newblock In \emph{Proceedings of the IEEE/CVF Conference on Computer Vision and Pattern Recognition}, pages 12824--12833, 2021.

\bibitem[Fuoli et~al.(2019)Fuoli, Gu, and Timofte]{fuoli2019efficient}
Dario Fuoli, Shuhang Gu, and Radu Timofte.
\newblock Efficient video super-resolution through recurrent latent space propagation.
\newblock In \emph{2019 IEEE/CVF International Conference on Computer Vision Workshop (ICCVW)}, pages 3476--3485. IEEE, 2019.

\bibitem[Gallego et~al.(2020)Gallego, Delbr{\"u}ck, Orchard, Bartolozzi, Taba, Censi, Leutenegger, Davison, Conradt, Daniilidis, et~al.]{gallego2020event}
Guillermo Gallego, Tobi Delbr{\"u}ck, Garrick Orchard, Chiara Bartolozzi, Brian Taba, Andrea Censi, Stefan Leutenegger, Andrew~J Davison, J{\"o}rg Conradt, Kostas Daniilidis, et~al.
\newblock Event-based vision: A survey.
\newblock \emph{IEEE transactions on pattern analysis and machine intelligence}, 44\penalty0 (1):\penalty0 154--180, 2020.

\bibitem[Gankhuyag et~al.(2023)Gankhuyag, Yoon, Park, Son, and Min]{gankhuyag2023lightweight}
Ganzorig Gankhuyag, Kihwan Yoon, Jinman Park, Haeng~Seon Son, and Kyoungwon Min.
\newblock Lightweight real-time image super-resolution network for 4k images.
\newblock In \emph{Proceedings of the IEEE/CVF Conference on Computer Vision and Pattern Recognition}, pages 1746--1755, 2023.

\bibitem[Gehrig and Scaramuzza(2022)]{gehrig2022high}
Daniel Gehrig and Davide Scaramuzza.
\newblock Are high-resolution event cameras really needed?
\newblock \emph{arXiv preprint arXiv:2203.14672}, 2022.

\bibitem[Gehrig et~al.(2019)Gehrig, Loquercio, Derpanis, and Scaramuzza]{gehrig2019end}
Daniel Gehrig, Antonio Loquercio, Konstantinos~G Derpanis, and Davide Scaramuzza.
\newblock End-to-end learning of representations for asynchronous event-based data.
\newblock In \emph{Proceedings of the IEEE/CVF International Conference on Computer Vision}, pages 5633--5643, 2019.

\bibitem[Geng et~al.(2022)Geng, Liang, Ding, and Zharkov]{geng2022rstt}
Zhicheng Geng, Luming Liang, Tianyu Ding, and Ilya Zharkov.
\newblock Rstt: Real-time spatial temporal transformer for space-time video super-resolution.
\newblock In \emph{Proceedings of the IEEE/CVF Conference on Computer Vision and Pattern Recognition}, pages 17441--17451, 2022.

\bibitem[Isobe et~al.(2022)Isobe, Jia, Tao, Li, Li, Shi, Mu, Lu, and Tai]{isobe2022look}
Takashi Isobe, Xu Jia, Xin Tao, Changlin Li, Ruihuang Li, Yongjie Shi, Jing Mu, Huchuan Lu, and Yu-Wing Tai.
\newblock Look back and forth: Video super-resolution with explicit temporal difference modeling.
\newblock In \emph{Proceedings of the IEEE/CVF Conference on Computer Vision and Pattern Recognition}, pages 17411--17420, 2022.

\bibitem[Kiani~Galoogahi et~al.(2017)Kiani~Galoogahi, Fagg, Huang, Ramanan, and Lucey]{kiani2017need}
Hamed Kiani~Galoogahi, Ashton Fagg, Chen Huang, Deva Ramanan, and Simon Lucey.
\newblock Need for speed: A benchmark for higher frame rate object tracking.
\newblock In \emph{Proceedings of the IEEE International Conference on Computer Vision}, pages 1125--1134, 2017.

\bibitem[Kim et~al.(2021)Kim, Bae, Park, Zhang, and Kim]{kim2021n}
Junho Kim, Jaehyeok Bae, Gangin Park, Dongsu Zhang, and Young~Min Kim.
\newblock N-imagenet: Towards robust, fine-grained object recognition with event cameras.
\newblock In \emph{Proceedings of the IEEE/CVF international conference on computer vision}, pages 2146--2156, 2021.

\bibitem[Li et~al.(2023{\natexlab{a}})Li, Li, and Tian]{li2023sodformer}
Dianze Li, Jianing Li, and Yonghong Tian.
\newblock Sodformer: Streaming object detection with transformer using events and frames.
\newblock \emph{IEEE Transactions on Pattern Analysis and Machine Intelligence}, 2023{\natexlab{a}}.

\bibitem[Li et~al.(2019{\natexlab{a}})Li, Li, and Shi]{li2019super}
Hongmin Li, Guoqi Li, and Luping Shi.
\newblock Super-resolution of spatiotemporal event-stream image.
\newblock \emph{Neurocomputing}, 335:\penalty0 206--214, 2019{\natexlab{a}}.

\bibitem[Li et~al.(2021)Li, Feng, Li, Jiang, Zou, and Gao]{li2021event}
Siqi Li, Yutong Feng, Yipeng Li, Yu Jiang, Changqing Zou, and Yue Gao.
\newblock Event stream super-resolution via spatiotemporal constraint learning.
\newblock In \emph{Proceedings of the IEEE/CVF International Conference on Computer Vision}, pages 4480--4489, 2021.

\bibitem[Li et~al.(2023{\natexlab{b}})Li, Zuo, and Loy]{li2023learning}
Xiaoming Li, Wangmeng Zuo, and Chen~Change Loy.
\newblock Learning generative structure prior for blind text image super-resolution.
\newblock In \emph{Proceedings of the IEEE/CVF Conference on Computer Vision and Pattern Recognition}, pages 10103--10113, 2023{\natexlab{b}}.

\bibitem[Li et~al.(2019{\natexlab{b}})Li, Yang, Liu, Yang, Jeon, and Wu]{li2019feedback}
Zhen Li, Jinglei Yang, Zheng Liu, Xiaomin Yang, Gwanggil Jeon, and Wei Wu.
\newblock Feedback network for image super-resolution.
\newblock In \emph{Proceedings of the IEEE/CVF conference on computer vision and pattern recognition}, pages 3867--3876, 2019{\natexlab{b}}.

\bibitem[Liang et~al.(2023)Liang, Zheng, Huang, Zhang, Chen, and Tian]{liang2023event}
Quanmin Liang, Xiawu Zheng, Kai Huang, Yan Zhang, Jie Chen, and Yonghong Tian.
\newblock Event-diffusion: Event-based image reconstruction and restoration with diffusion models.
\newblock In \emph{Proceedings of the 31st ACM International Conference on Multimedia}, pages 3837--3846, 2023.

\bibitem[Lichtsteiner et~al.(2008)Lichtsteiner, Posch, and Delbr{\"{u}}ck]{lichtsteiner2008128}
Patrick Lichtsteiner, Christoph Posch, and Tobi Delbr{\"{u}}ck.
\newblock A 128{\texttimes}128 120 db 15 {\(\mathrm{\mu}\)}s latency asynchronous temporal contrast vision sensor.
\newblock \emph{{IEEE} J. Solid State Circuits}, 43\penalty0 (2):\penalty0 566--576, 2008.

\bibitem[Lin et~al.(2022)Lin, Ma, Guo, and Wen]{lin2022dvs}
Songnan Lin, Ye Ma, Zhenhua Guo, and Bihan Wen.
\newblock Dvs-voltmeter: Stochastic process-based event simulator for dynamic vision sensors.
\newblock In \emph{European Conference on Computer Vision}, pages 578--593. Springer, 2022.

\bibitem[Lu et~al.(2023)Lu, Wang, Liu, Wang, and Wang]{lu2023learning}
Yunfan Lu, Zipeng Wang, Minjie Liu, Hongjian Wang, and Lin Wang.
\newblock Learning spatial-temporal implicit neural representations for event-guided video super-resolution.
\newblock In \emph{Proceedings of the IEEE/CVF Conference on Computer Vision and Pattern Recognition}, pages 1557--1567, 2023.

\bibitem[Maqueda et~al.(2018)Maqueda, Loquercio, Gallego, Garc{\'\i}a, and Scaramuzza]{maqueda2018event}
Ana~I Maqueda, Antonio Loquercio, Guillermo Gallego, Narciso Garc{\'\i}a, and Davide Scaramuzza.
\newblock Event-based vision meets deep learning on steering prediction for self-driving cars.
\newblock In \emph{Proceedings of the IEEE conference on computer vision and pattern recognition}, pages 5419--5427, 2018.

\bibitem[Messikommer et~al.(2023)Messikommer, Fang, Gehrig, and Scaramuzza]{messikommer2023data}
Nico Messikommer, Carter Fang, Mathias Gehrig, and Davide Scaramuzza.
\newblock Data-driven feature tracking for event cameras.
\newblock In \emph{Proceedings of the IEEE/CVF Conference on Computer Vision and Pattern Recognition}, pages 5642--5651, 2023.

\bibitem[Rebecq et~al.(2019)Rebecq, Ranftl, Koltun, and Scaramuzza]{rebecq2019high}
Henri Rebecq, Ren{\'e} Ranftl, Vladlen Koltun, and Davide Scaramuzza.
\newblock High speed and high dynamic range video with an event camera.
\newblock \emph{IEEE transactions on pattern analysis and machine intelligence}, 43\penalty0 (6):\penalty0 1964--1980, 2019.

\bibitem[Shi et~al.(2016)Shi, Caballero, Husz{\'a}r, Totz, Aitken, Bishop, Rueckert, and Wang]{shi2016real}
Wenzhe Shi, Jose Caballero, Ferenc Husz{\'a}r, Johannes Totz, Andrew~P Aitken, Rob Bishop, Daniel Rueckert, and Zehan Wang.
\newblock Real-time single image and video super-resolution using an efficient sub-pixel convolutional neural network.
\newblock In \emph{Proceedings of the IEEE conference on computer vision and pattern recognition}, pages 1874--1883, 2016.

\bibitem[Sironi et~al.(2018)Sironi, Brambilla, Bourdis, Lagorce, and Benosman]{sironi2018hats}
Amos Sironi, Manuele Brambilla, Nicolas Bourdis, Xavier Lagorce, and Ryad Benosman.
\newblock Hats: Histograms of averaged time surfaces for robust event-based object classification.
\newblock In \emph{Proceedings of the IEEE conference on computer vision and pattern recognition}, pages 1731--1740, 2018.

\bibitem[Stoffregen et~al.(2020)Stoffregen, Scheerlinck, Scaramuzza, Drummond, Barnes, Kleeman, and Mahony]{stoffregen2020reducing}
Timo Stoffregen, Cedric Scheerlinck, Davide Scaramuzza, Tom Drummond, Nick Barnes, Lindsay Kleeman, and Robert Mahony.
\newblock Reducing the sim-to-real gap for event cameras.
\newblock In \emph{Computer Vision--ECCV 2020: 16th European Conference, Glasgow, UK, August 23--28, 2020, Proceedings, Part XXVII 16}, pages 534--549. Springer, 2020.

\bibitem[Tan et~al.(2022)Tan, Wang, Han, Cao, Wu, and Zha]{tan2022multi}
Ganchao Tan, Yang Wang, Han Han, Yang Cao, Feng Wu, and Zheng-Jun Zha.
\newblock Multi-grained spatio-temporal features perceived network for event-based lip-reading.
\newblock In \emph{Proceedings of the IEEE/CVF Conference on Computer Vision and Pattern Recognition}, pages 20094--20103, 2022.

\bibitem[Taverni et~al.(2018)Taverni, Moeys, Li, Cavaco, Motsnyi, Bello, and Delbruck]{taverni2018front}
Gemma Taverni, Diederik~Paul Moeys, Chenghan Li, Celso Cavaco, Vasyl Motsnyi, David San~Segundo Bello, and Tobi Delbruck.
\newblock Front and back illuminated dynamic and active pixel vision sensors comparison.
\newblock \emph{IEEE Transactions on Circuits and Systems II: Express Briefs}, 65\penalty0 (5):\penalty0 677--681, 2018.

\bibitem[Tulyakov et~al.(2021)Tulyakov, Gehrig, Georgoulis, Erbach, Gehrig, Li, and Scaramuzza]{tulyakov2021time}
Stepan Tulyakov, Daniel Gehrig, Stamatios Georgoulis, Julius Erbach, Mathias Gehrig, Yuanyou Li, and Davide Scaramuzza.
\newblock Time lens: Event-based video frame interpolation.
\newblock In \emph{Proceedings of the IEEE/CVF conference on computer vision and pattern recognition}, pages 16155--16164, 2021.

\bibitem[Tulyakov et~al.(2022)Tulyakov, Bochicchio, Gehrig, Georgoulis, Li, and Scaramuzza]{tulyakov2022time}
Stepan Tulyakov, Alfredo Bochicchio, Daniel Gehrig, Stamatios Georgoulis, Yuanyou Li, and Davide Scaramuzza.
\newblock Time lens++: Event-based frame interpolation with parametric non-linear flow and multi-scale fusion.
\newblock In \emph{Proceedings of the IEEE/CVF Conference on Computer Vision and Pattern Recognition}, pages 17755--17764, 2022.

\bibitem[Vaswani et~al.(2017)Vaswani, Shazeer, Parmar, Uszkoreit, Jones, Gomez, Kaiser, and Polosukhin]{vaswani2017attention}
Ashish Vaswani, Noam Shazeer, Niki Parmar, Jakob Uszkoreit, Llion Jones, Aidan~N Gomez, {\L}ukasz Kaiser, and Illia Polosukhin.
\newblock Attention is all you need.
\newblock \emph{Advances in neural information processing systems}, 30, 2017.

\bibitem[Wang et~al.(2023)Wang, Isobe, Jia, Tao, Lu, and Tai]{wang2023compression}
Yingwei Wang, Takashi Isobe, Xu Jia, Xin Tao, Huchuan Lu, and Yu-Wing Tai.
\newblock Compression-aware video super-resolution.
\newblock In \emph{Proceedings of the IEEE/CVF Conference on Computer Vision and Pattern Recognition}, pages 2012--2021, 2023.

\bibitem[Wang et~al.(2004)Wang, Bovik, Sheikh, and Simoncelli]{wang2004image}
Zhou Wang, Alan~C. Bovik, Hamid~R. Sheikh, and Eero~P. Simoncelli.
\newblock Image quality assessment: from error visibility to structural similarity.
\newblock \emph{{IEEE} Trans. Image Process.}, 13\penalty0 (4):\penalty0 600--612, 2004.

\bibitem[Wang et~al.(2020)Wang, Duan, Cossairt, Katsaggelos, Huang, and Shi]{wang2020joint}
Zihao~W Wang, Peiqi Duan, Oliver Cossairt, Aggelos Katsaggelos, Tiejun Huang, and Boxin Shi.
\newblock Joint filtering of intensity images and neuromorphic events for high-resolution noise-robust imaging.
\newblock In \emph{Proceedings of the IEEE/CVF Conference on Computer Vision and Pattern Recognition}, pages 1609--1619, 2020.

\bibitem[Weng et~al.(2021)Weng, Zhang, and Xiong]{weng2021event}
Wenming Weng, Yueyi Zhang, and Zhiwei Xiong.
\newblock Event-based video reconstruction using transformer.
\newblock In \emph{2021 {IEEE/CVF} International Conference on Computer Vision, {ICCV} 2021, Montreal, QC, Canada, October 10-17, 2021}, pages 2543--2552. {IEEE}, 2021.

\bibitem[Weng et~al.(2022)Weng, Zhang, and Xiong]{weng2022boosting}
Wenming Weng, Yueyi Zhang, and Zhiwei Xiong.
\newblock Boosting event stream super-resolution with a recurrent neural network.
\newblock In \emph{European Conference on Computer Vision}, pages 470--488. Springer, 2022.

\bibitem[Xu et~al.(2020)Xu, Xu, Golyanik, Habermann, Fang, and Theobalt]{xu2020eventcap}
Lan Xu, Weipeng Xu, Vladislav Golyanik, Marc Habermann, Lu Fang, and Christian Theobalt.
\newblock Eventcap: Monocular 3d capture of high-speed human motions using an event camera.
\newblock In \emph{Proceedings of the IEEE/CVF Conference on Computer Vision and Pattern Recognition}, pages 4968--4978, 2020.

\bibitem[Zhang et~al.(2018)Zhang, Isola, Efros, Shechtman, and Wang]{zhang2018unreasonable}
Richard Zhang, Phillip Isola, Alexei~A. Efros, Eli Shechtman, and Oliver Wang.
\newblock The unreasonable effectiveness of deep features as a perceptual metric.
\newblock In \emph{2018 {IEEE} Conference on Computer Vision and Pattern Recognition, {CVPR} 2018, Salt Lake City, UT, USA, June 18-22, 2018}, pages 586--595. Computer Vision Foundation / {IEEE} Computer Society, 2018.

\bibitem[Zhou et~al.(2018)Zhou, Gallego, Rebecq, Kneip, Li, and Scaramuzza]{zhou2018semi}
Yi Zhou, Guillermo Gallego, Henri Rebecq, Laurent Kneip, Hongdong Li, and Davide Scaramuzza.
\newblock Semi-dense 3d reconstruction with a stereo event camera.
\newblock In \emph{Proceedings of the European conference on computer vision (ECCV)}, pages 235--251, 2018.

\bibitem[Zhu et~al.(2018)Zhu, Yuan, Chaney, and Daniilidis]{zhu2018ev}
Alex~Zihao Zhu, Liangzhe Yuan, Kenneth Chaney, and Kostas Daniilidis.
\newblock Ev-flownet: Self-supervised optical flow estimation for event-based cameras.
\newblock \emph{arXiv preprint arXiv:1802.06898}, 2018.

\end{thebibliography}
}
\clearpage
\appendix
\section{Dataset and Training Configuration}
\label{app-config}
\textbf{Synthetic Data Generation Method:} We synthesized the NFS-syn and RGB-syn datasets using an event simulator \cite{lin2022dvs}. Specifically, we initially downsampled the NFS dataset \cite{kiani2017need} and RGB-DAVIS dataset \cite{wang2020joint} using bicubic interpolation to obtain low-resolution images. The original resolution of the NFS dataset is \(1280\times720\), and we applied \(2(4,8,16)\times\) downsampling, while the RGB-DAVIS dataset has an original resolution of \(1520\times1440\), and we downsampled it by \(2(4,8)\times\). Subsequently, we generated event streams through the event simulator, utilizing default initial parameters. In the NFS-syn dataset, we considered the \(16\times\) downsampled \(80\times45\) data as the minimum resolution and the \(2\times\) downsampled \(360\times180\) data as the maximum resolution, resulting in LR-HR pairs at \(2(4,8)\times\). For the RGB-syn dataset, we used the \(8\times\) downsampled \(190\times180\) data as the minimum resolution and the \(2\times\) downsampled \(760\times720\) data as the maximum resolution, resulting in LR-HR pairs at \(2(4)\times\).
And we applied data augmentations such as random flipping and polarity inversion to the dataset, and randomly split it for training and testing. 

\noindent\textbf{Training Configuration:} During training, we applied several enhancements to the event stream data, including vertical and horizontal flipping of event count images and flipping event stream polarities, each with a probability of 50\%. We randomly partitioned the data into training and testing sets for EventNFS, NFS-syn, and RGB-syn. Training and testing were performed on each dataset.

\section{Pseudo Code of \module}
\label{pseudo-code}

We provide the pseudo code of \module, as presented in \cref{pseudocode}.

\begin{algorithm}[t]
        \caption{Pseudo code of \module} 
        	\label{alg:algorithm} 
        \begin{lstlisting}[language=Python]
# xp, xn: features of positive, negative event, respectively
# xint: the cross-layer interaction representation

def BIE(xp, xn, xint):
    b, c, h, w = xp.shape
    scale_factor = c ** 0.5
    
    # (1) update xp and xn
    xp_ = convp(xp)
    xn_ = convn(xn)

    # (2) obtain the query Q and the value V
    Qp = conv_qp(LayerNorm(Conv2D(torch.cat([xint, xp],dim=1)))).view(b,c,-1) 
    Qn = conv_qn(LayerNorm(Conv2D(torch.cat([xint, xn],dim=1)))).view(b,c,-1) 
    Vp = conv_vp(xp).view(b,c,-1)
    Vn = conv_vn(xn).view(b,c,-1)

    Kp = Vp.permute(0,2,1)
    Kn = Vn.permute(0,2,1)
    
    # (3) calculate attention scores
    A_n2p = torch.bmm(Qp, Kn) * scale_factor
    A_p2n = torch.bmm(Qn, Kp) * scale_factor

    # (4) information propagation
    x_n2p = torch.bmm(torch.softmax(A_n2p,dim=-1), Vn).view(b,c,h,w)
    x_p2n = torch.bmm(torch.softmax(A_p2n,dim=-1), Vp).view(b,c,h,w)

    # (5) gated information fusion
    wp = torch.sigmoid(conv_wp1(xp_) + conv_wp2(x_n2p))  
    xp = wp*xp_ + (1-wp)*x_n2p
    wn = torch.sigmoid(conv_wn1(xn_) + conv_wn2(x_p2n))  
    xn = wn*xn_ + (1-wn)*x_p2n

    # (6) update xint
    xint = conv_int(torch.cat([Qp.view(b,c,h,w), Qn.view(b,c,h,w)], dim=1)) + xint

    return xp, xn, xint
\end{lstlisting}
\label{pseudocode}
\end{algorithm}

\section{More Ablation Study}

\subsection{Effect of CIR}
\begin{figure}[ht]
  \centering
  \includegraphics[width=0.47\textwidth]{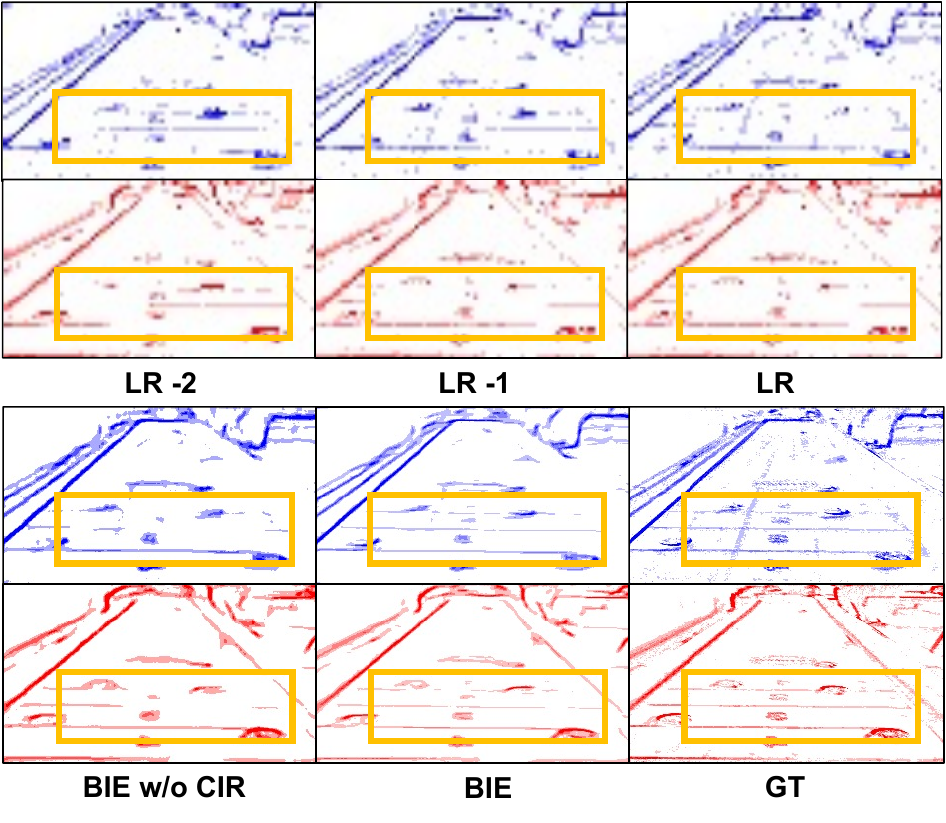}
  \caption{Qualitative comparison between \module equipped with and without CIR.}
  \label{fig:abla-rec}
\end{figure}

To verify the effectiveness of CIR in our proposed \module, we simply remove the CIR in \module to make a qualitative comparison with \module (for the quantitative comparisons, please refer to Tab. \textcolor{red}{3} in the \textbf{main paper}.). Moreover, we present the previous LR frames which are denoted as $LR -i$, $i=1,2$.
As shown in \cref{fig:abla-rec}, the introduction of CIR in the \module enables our model to more effectively capture contextual information from different events in previous frames and utilize information that does not exist in the current frame, thereby enhancing the performance.

\subsection{Hyperparameters of Architecture}
We conducted experiments to explore the impact of hyperparameters on the super-resolution results of the \method, specifically focusing on two aspects: 1) ablation of layers in the residual network module, and 2) ablation of the channel size in feature extraction. To investigate the influence of model hyperparameters on the super-resolution outcomes, we modified the model parameters and retrained them on the NFS-syn dataset. The \(4\times\) SR results on both the NFS-syn dataset are presented in \cref{tab:hyper}. Additionally, we report their model parameters and FLOPs for model and time complexity analysis. 
The results show that adjusting the number of layers has a minor impact on the performance of \method, but it significantly reduces FLOPs. On the other hand, reducing the channel size substantially decreases the parameter count while significantly affecting the performance. After considering the trade-off between efficiency and effectiveness, we have set $C$=128 and $N$=5 in practice.

\begin{table}[h]
\centering
\begin{adjustbox}{width=0.8\linewidth}
    \renewcommand{\arraystretch}{1.2}
\begin{tabular}{l|ccc}
\toprule
($C$, $N$) & NFS-syn  & \# Params & \# FLOPs \\
\midrule
(32, 5) & 0.598 & \textbf{0.2} M & \textbf{2.2} G \\
(64, 5) & 0.571 & \underline{0.7} M & \underline{8.9} G \\
(128, 5) & \underline{0.552} & 2.7 M & 35.7 G \\
(256, 5) & 0.553 & 10.6 M & 141.7 G \\
(128, 3) & 0.554 & 2.5 M & 22.9 G \\
(128, 10) & \textbf{0.547} & 3.2 M & 66.3 G \\
\bottomrule
\end{tabular}
\renewcommand{\arraystretch}{1}
    \end{adjustbox}
\caption{The effect of channel $C$ and the number of basic blocks $N$ in \method. The FLOPs is calculated on the LR events of NFS-syn dataset with resolutions of 80$\times$45.}
\vspace{-3mm}
\label{tab:hyper}
\end{table}

\noindent\textbf{Effectiveness of The Number of Global Structures in \module.}
To investigate the effect of the number of global structures $M$ in \module, we conduct a series experiments by setting different $M$. The channel $C$=128, resolution $(H, W)$ of LR events is $(80, 45)$, and the scale $S$=4. The results are present in \cref{tab:ncluster}. In practice, we set $M$=128 from the consideration of the trade-off between efficiency and performance.
\begin{table}[h]
\centering
\begin{adjustbox}{width=0.7\linewidth}
    \renewcommand{\arraystretch}{1.2}
\begin{tabular}{lccc}
\toprule
Methods & NFS-syn & \# Params & \# FLOPs \\
\midrule
$M$=8 & 0.581 & \textbf{335.6} K & \textbf{1.19} G \\
$M$=16 & 0.580 & \underline{342.8} K & \underline{1.21} G \\
$M$=32 & 0.578 & 357.2 K & 1.24 G \\
$M$=64 & 0.579 & 385.9 K & 1.30 G \\
$M$=128 & 0.577 & 443.5 K & 1.42 G \\
$M$=256 & \underline{0.576} & 558.5 K & 1.50 G \\
$M$=512 & \textbf{0.574} & 778.7 K & 2.13 G \\
\bottomrule
\end{tabular}
\renewcommand{\arraystretch}{1}
    \end{adjustbox}
\caption{The effect of the number of global structures in \module. The experiments are conducted on NFS-syn dataset for 4$\times$ SR on RMSE metrics}
\vspace{-3mm}
\label{tab:ncluster}
\end{table}

\subsection{The Generalization in Real Event Dataset}
\label{generalization}
To validate the generalization of our models, we applied the models trained on the synthetic dataset NFS-syn to perform super-resolution on the real dataset EventNFS \cite{duan2021eventzoom}. From \cref{tab:generalization}, it can be observed that although our methods \method-plain and \method still outperform other approaches, this advantage is somewhat diminished. There could be two reasons for this. First, the minimum resolution of EventNFS is \(55\times31\), which is too small, causing significant degradation of event information and hindering the effective recovery of high-resolution event streams. Second, differences between real and synthetic event streams may result in a weakened performance of the model. Investigating methods to mitigate the impact of differences between real and synthetic event streams is an important direction for our future research.

\begin{table}[t]
\centering
\begin{adjustbox}{width=0.3\textwidth}
    \renewcommand{\arraystretch}{1.2}
\begin{tabular}{lll}
\toprule
\multirow{2}{*}{Methods} & \multicolumn{2}{c}{EventNFS-real*} \\ \cmidrule{2-3} 
                         & \multicolumn{1}{c}{\(2\times\)}               & \multicolumn{1}{c}{\(4\times\)}              \\ \midrule
bicubic                  & 0.872            & 0.948           \\
SRFBN \cite{li2019feedback}                    & 0.837            & 0.901           \\
RLSP \cite{fuoli2019efficient}                     & 0.837            & 0.917           \\
RSTT \cite{geng2022rstt}                     & 0.812            & 0.887           \\
EventZoom \cite{duan2021eventzoom}                & 1.043            & 1.117           \\
RecEvSR \cite{weng2022boosting}                  & 0.822            & 0.897           \\
BMCNet-plain             & \underline{0.804}            & \textbf{0.869}           \\
BMCNet                   & \textbf{0.792}            & \underline{0.877}           \\ \bottomrule
\end{tabular}
\renewcommand{\arraystretch}{1}
    \end{adjustbox}
\caption{Quantitative analysis results for super-resolution on EventNFS-real dataset using bicubic, SRFBN \cite{li2019feedback}, RLSP \cite{fuoli2019efficient}, RSTT \cite{geng2022rstt}, EventZoom \cite{duan2021eventzoom}, RecEvSR \cite{weng2022boosting}, and our methods \method-plain and \method. * denotes that all models are trained on the synthetic dataset NFS-syn and then applied to super-resolution on the real dataset EventNFS-real. We report RMSE results for each super-resolution scale.}
\label{tab:generalization}
\end{table}

\section{More Visual Results}
\subsection{More Qualitative Comparison Results}
\label{app-more-quali-results}
\begin{figure*}[ht]
  \centering
  \includegraphics[width=\textwidth]{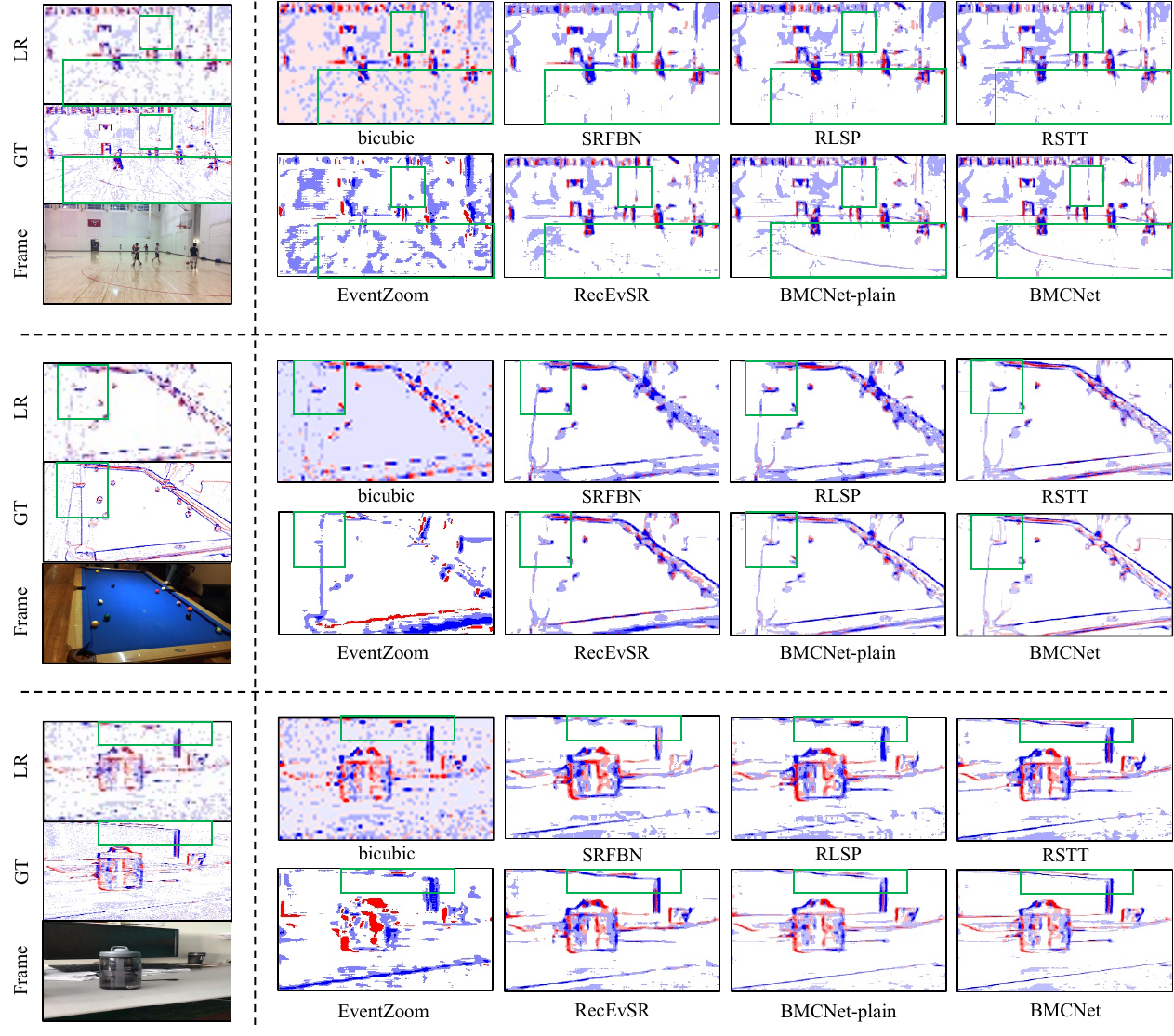}
  \caption{Super-Resolution Results at \(4\times\) Scale on the NFS-syn Dataset for bicubic, SRFBN \cite{li2019feedback}, RLSP \cite{fuoli2019efficient}, RSTT \cite{geng2022rstt}, EventZoom \cite{duan2021eventzoom}, RecEvSR \cite{weng2022boosting}, and our methods \method-plain and \method. \textbf{[Best viewed with zoom-in.]}}
  \label{nfs}
\end{figure*}

\begin{figure*}[ht]
  \centering
  \includegraphics[width=0.91\textwidth]{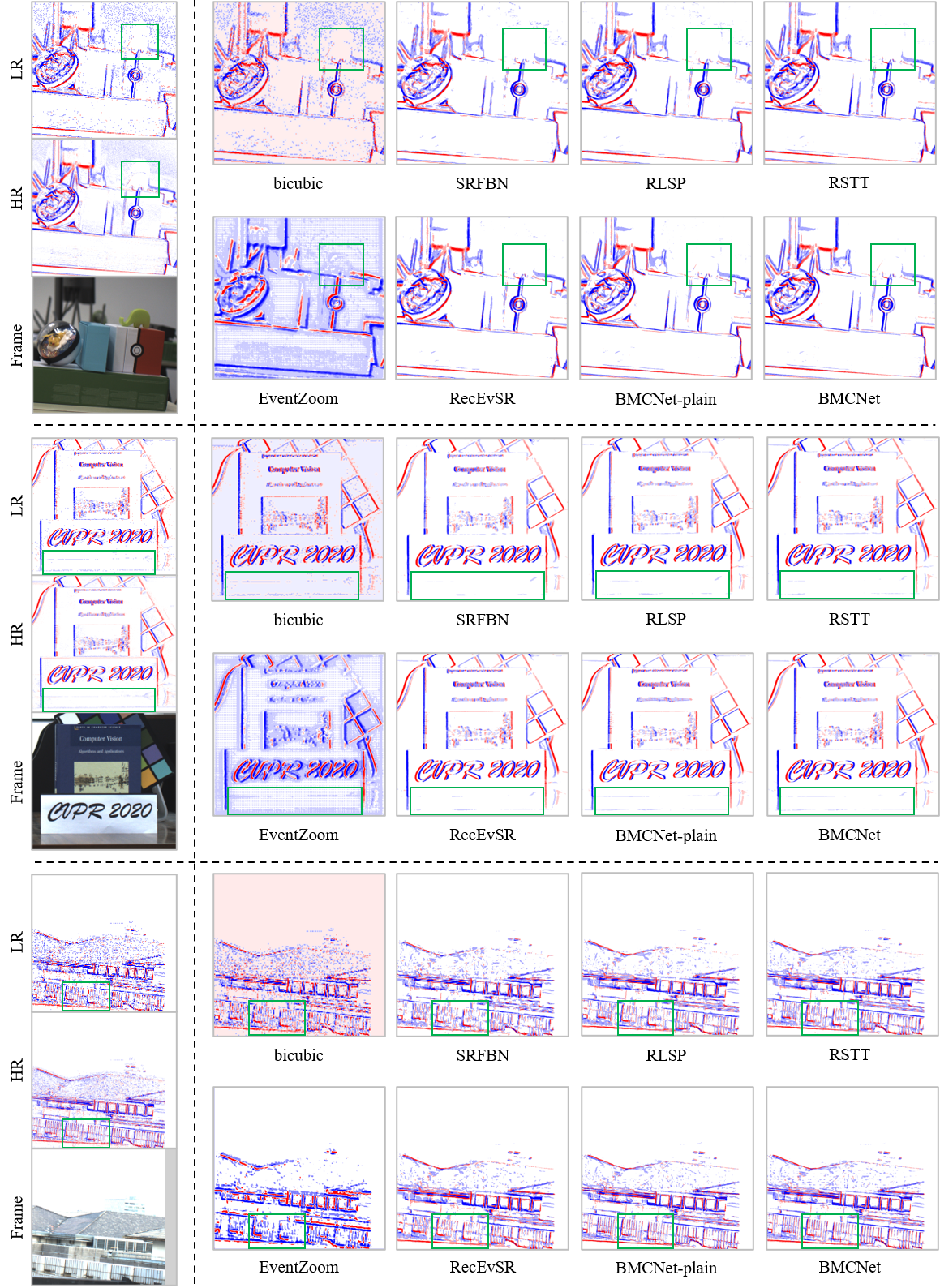}
  \caption{Super-Resolution Results at \(4\times\) Scale on the RGB-syn Dataset for bicubic, SRFBN \cite{li2019feedback}, RLSP \cite{fuoli2019efficient}, RSTT \cite{geng2022rstt}, EventZoom \cite{duan2021eventzoom}, RecEvSR \cite{weng2022boosting}, and our methods \method-plain and \method. \textbf{[Best viewed with zoom-in.]}}
  \label{RGB}
\end{figure*}

\begin{figure*}[ht]
  \centering
  \includegraphics[width=\textwidth]{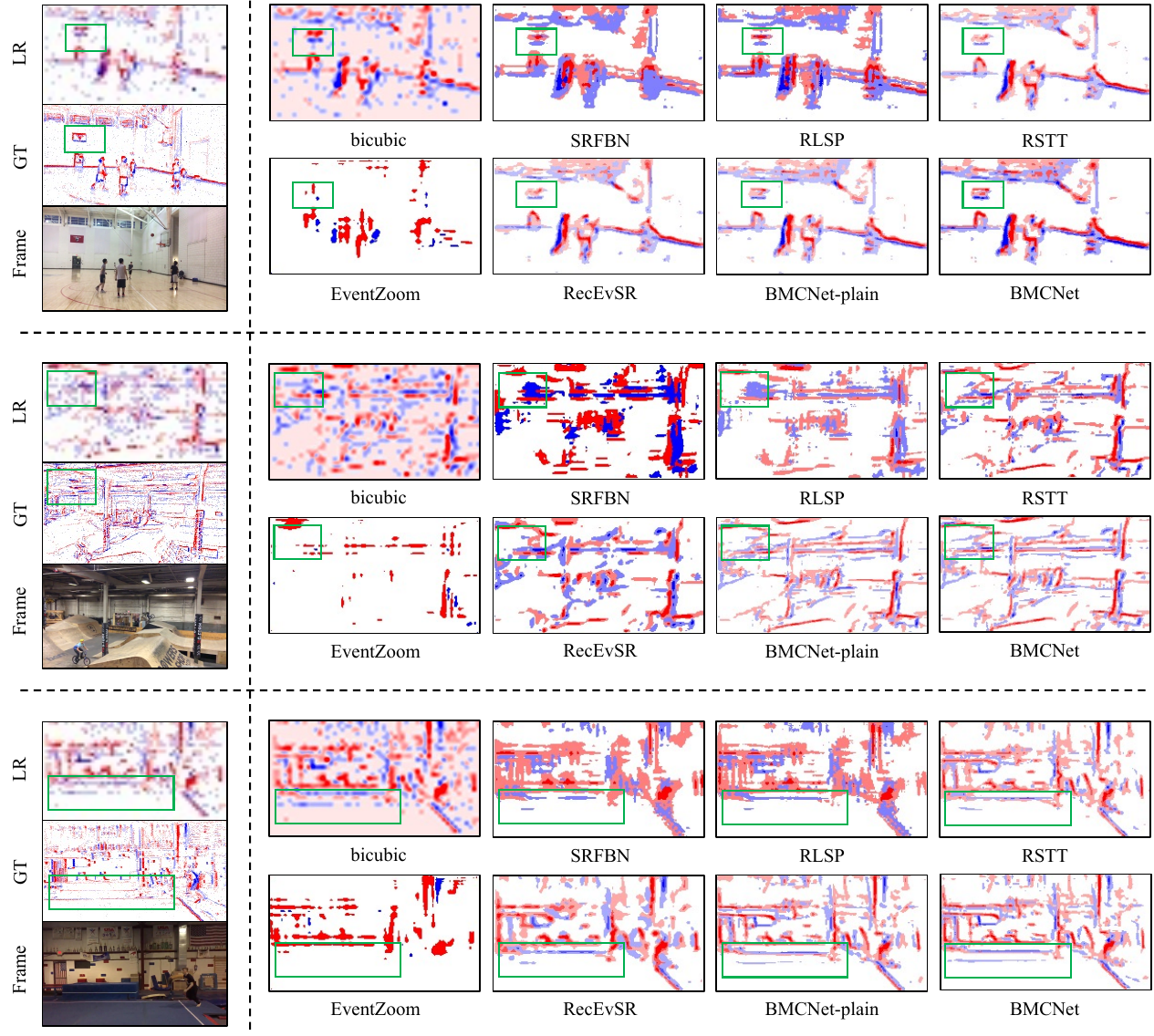}
  \caption{Super-Resolution Results at \(4\times\) Scale on the EventNFS-real Dataset for bicubic, SRFBN \cite{li2019feedback}, RLSP \cite{fuoli2019efficient}, RSTT \cite{geng2022rstt}, EventZoom \cite{duan2021eventzoom}, RecEvSR \cite{weng2022boosting}, and our methods \method-plain and \method. \textbf{[Best viewed with zoom-in.]}}
  \label{real}
\end{figure*}

In \cref{nfs}, \cref{RGB}, and \cref{real}, we present the \(4\times\) super-resolution results on the NFS-syn, RGB-syn, and EventNFS-real datasets for bicubic, SRFBN \cite{li2019feedback}, RLSP \cite{fuoli2019efficient}, RSTT \cite{geng2022rstt}, EventZoom \cite{duan2021eventzoom}, RecEvSR \cite{weng2022boosting}, and our methods \method-plain and \method. The low-resolution (LR) resolutions for NFS-syn, RGB-syn, and EventNFS-real are \(80\times45\), \(190\times180\), and \(55\times31\), respectively.  It is evident from the results that our \method-plain and \method excel in integrating overall structural information to complement and rectify detailed information, resulting in richer details and clearer edges.

\subsection{Event-Based Video Reconstruction}
\label{app-more-downstream-results}
\begin{figure*}[ht]
  \centering
  \includegraphics[width=0.96\textwidth]{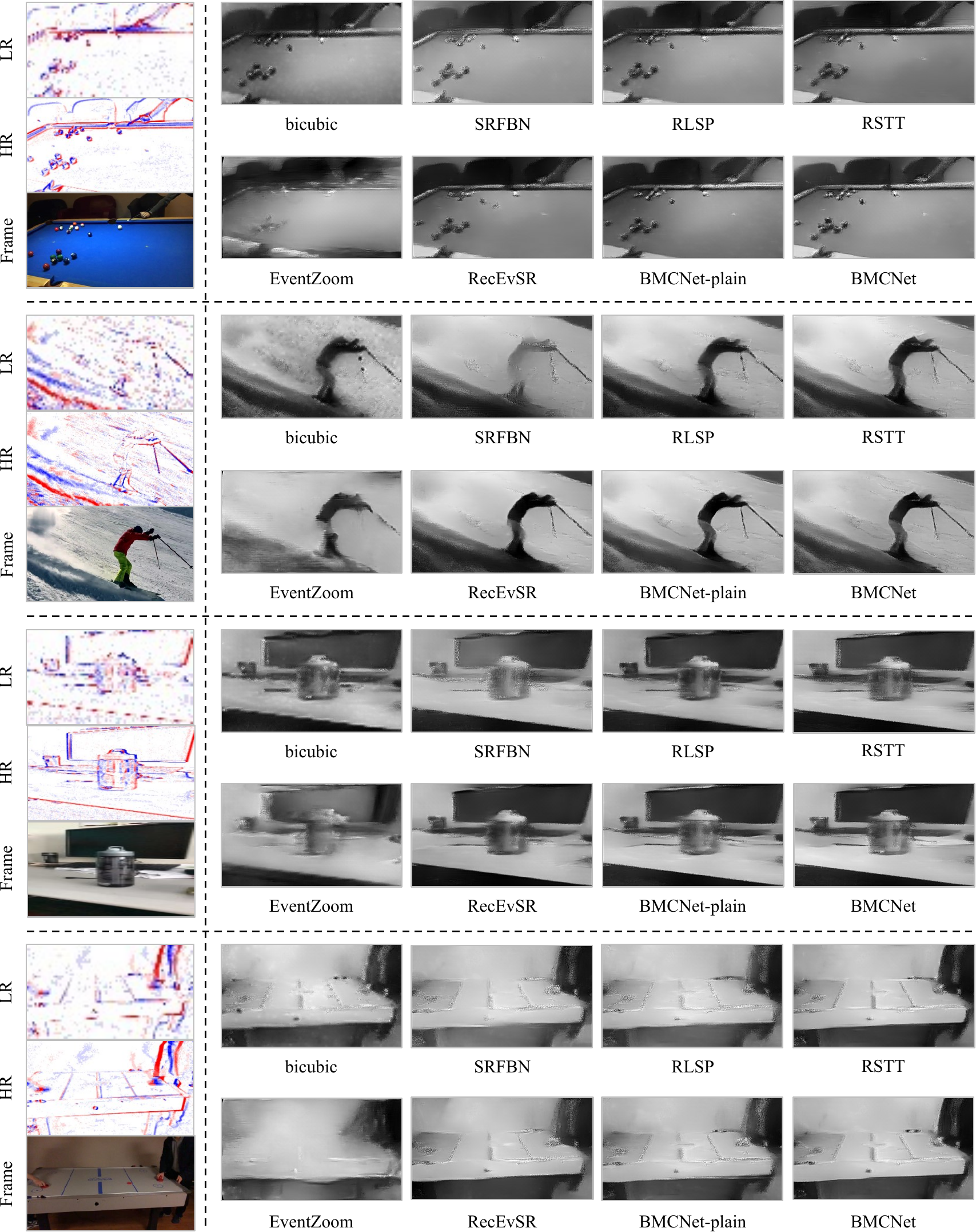}
  \caption{Qualitative analysis results for event-based video reconstruction comparing bicubic, SRFBN \cite{li2019feedback}, RLSP \cite{fuoli2019efficient}, RSTT \cite{geng2022rstt}, EventZoom \cite{duan2021eventzoom}, RecEvSR \cite{weng2022boosting}, and our methods \method-plain and \method. The results showcase the performance of all methods in \(4\times\) super-resolution on the NFS-syn dataset. \textbf{[Best viewed with zoom-in.]}}
  \label{reconstruction}
\end{figure*}

As shown in \cref{reconstruction}, we applied bicubic, SRFBN \cite{li2019feedback}, RLSP \cite{fuoli2019efficient}, RSTT \cite{geng2022rstt}, EventZoom \cite{duan2021eventzoom}, RecEvSR \cite{weng2022boosting}, and our methods \method-plain and \method to perform \(4\times\) super-resolution on the downsampled NFS-syn dataset. Subsequently, we utilized the E2VID \cite{rebecq2019high} to reconstruct the events based on the super-resolved event stream. It is evident that our methods \method-plain and \method achieve superior reconstruction with finer details and fewer artifacts, validating the effectiveness of our approach.

\end{document}